\documentclass[english]{article}
\usepackage[T1]{fontenc}
\usepackage[utf8]{inputenc}
\usepackage{float}
\usepackage{amsmath}
\usepackage{amsthm}
\usepackage{amssymb}
\usepackage{graphicx}

\makeatletter

\floatstyle{ruled}
\newfloat{algorithm}{tbp}{loa}
\providecommand{\algorithmname}{Algorithm}
\floatname{algorithm}{\protect\algorithmname}

  \theoremstyle{plain}
  \newtheorem{assumption}{\protect\assumptionname}
\theoremstyle{plain}
\newtheorem{thm}{\protect\theoremname}
  \theoremstyle{remark}
  \newtheorem{rem}{\protect\remarkname}

\usepackage[nonatbib,final]{nips_2016}
\usepackage[numbers,compress]{natbib}

\usepackage{hyperref}       
\usepackage{url}            
\usepackage{booktabs}       
\usepackage{amsfonts}       
\usepackage{nicefrac}       
\usepackage{microtype}      

\usepackage{algorithmicx}

\algnewcommand\algorithmicinput{\textbf{INPUT:}}
\algnewcommand\INPUT{\item[\algorithmicinput]}
\algnewcommand\algorithmicoutput{\textbf{OUTPUT:}}
\algnewcommand\OUTPUT{\item[\algorithmicoutput]}
\algnewcommand\algorithmicparameter{\textbf{PARAMETER:}}
\algnewcommand\PARAMETER{\item[\algorithmicparameter]}

\usepackage{amsmath}

\usepackage[resetlabels,labeled]{multibib}
\newcites{app}{References}

\newtheoremstyle{mystyle}
  {}
  {}
  {\itshape}
  {}
  {\bfseries}
  {.}
  { }
  {}

\theoremstyle{mystyle}

\DeclareMathOperator*{\plim}{p
lim}

\usepackage{esvect}

\makeatother

\usepackage{babel}
  \providecommand{\assumptionname}{Assumption}
  \providecommand{\remarkname}{Remark}
\providecommand{\theoremname}{Theorem}

\begin{document}

\title{Spectral Learning of Dynamic Systems from Nonequilibrium Data}

\author{Hao Wu and Frank No\'e\\
Department of Mathematics and Computer Science\\
Freie Universität Berlin\\
Arnimallee 6, 14195 Berlin\\
\texttt{\{hao.wu,frank.noe\}@fu-berlin.de} \\}
\maketitle
\begin{abstract}
Observable operator models (OOMs) and related models are one of the
most important and powerful tools for modeling and analyzing stochastic
systems. They exactly describe dynamics of finite-rank systems and
can be efficiently and consistently estimated through spectral learning
under the assumption of identically distributed data. In this paper,
we investigate the properties of spectral learning without this assumption
due to the requirements of analyzing large-time scale systems, and
show that the equilibrium dynamics of a system can be extracted from
nonequilibrium observation data by imposing an equilibrium constraint.
In addition, we propose a binless extension of spectral learning for
continuous data. In comparison with the other continuous-valued spectral
algorithms, the binless algorithm can achieve consistent estimation
of equilibrium dynamics with only linear complexity.
\end{abstract}

\section{Introduction}

In the last two decades, a collection of highly related dynamic models
including observable operator models (OOMs) \cite{jaeger2000observable,zhao2009bound,Jaeger2012Discrete},
predictive state representations \cite{littman2001predictive,Singh2004predictive,wiewiora2005learning}
and reduced-rank hidden Markov models \cite{hsu2009spectral,siddiq2010reduced},
have become powerful and increasingly popular tools for analysis of
dynamic data. These models are largely similar, and all can be learned
by spectral methods in a general framework of multiplicity automata,
or equivalently sequential systems \cite{beimel2000learning,thon2015links}.
In contrast with the other commonly used models such as Markov state
models \cite{prinz2011markov,bowman2013introduction}, Langevin models
\cite{ruttor2013approximate,schaudinnus2015multidimensional}, traditional
hidden Markov models (HMMs) \cite{rabiner1989tutorial,Noe2013projected},
Gaussian process state-space models \cite{turner2010state,svensson2016computationally}
and recurrent neural networks \cite{hochreiter1997long}, the spectral
learning based models can exactly characterize the dynamics of a stochastic
system without any a priori knowledge except the assumption of finite
dynamic rank (i.e., the rank of Hankel matrix) \cite{thon2015links,wu2015projected},
and the parameter estimation can be efficiently performed for discrete-valued
systems without solving any intractable inverse or optimization problem.
We focus in this paper only on stochastic systems without control
inputs and all spectral learning based models can be expressed in
the form of OOMs for such systems, so we will refer to them as OOMs
below.

In most literature on spectral learning, the observation data are
assumed to be identically (possibly not independently) distributed
so that the expected values of observables associated with the parameter
estimation can be reliably computed by empirical averaging. However,
this assumption can be severely violated due to the limit of experimental
technique or computational capacity in many practical situations,
especially where metastable physical or chemical processes are involved.
A notable example is the distributed computing project Folding@home
\cite{shirts2000screen}, which explores protein folding processes
that occur on the timescales of microseconds to milliseconds based
on molecular dynamics simulations on the order of nanoseconds in length.
In such a nonequilibrium case where distributions of observation data
are time-varying and dependent on initial conditions, it is still
unclear if promising estimates of OOMs can be obtained. In \cite{huang2013spectral},
a hybrid estimation algorithm was proposed to improve spectral learning
of large-time scale processes by using both dynamic and static data,
but it still requires assumption of identically distributed data.
One solution to reduce the statistical bias caused by nonequilibrium
data is to discard the observation data generated before the system
reaches steady state, which is a common trick in applied statistics
\cite{cowles1996markov}. Obviously, this way suffers from substantial
information loss and is infeasible when observation trajectories are
shorter than mixing times. Another possible way would be to learn
OOMs by likelihood-based estimation instead of spectral methods, but
there is no effective maximum likelihood or Bayesian estimator of
OOMs until now. The maximum pseudo-likelihood estimator of OOMs proposed
in \cite{jiang2016improving} demands high computational cost and
its consistency is yet unverified.

Another difficulty for spectral approaches is learning with continuous
data, where density estimation problems are involved. The density
estimation can be performed by parametric methods such as the fuzzy
interpolation \cite{Jaeger2001Continuous} and the kernel density
estimation \cite{siddiq2010reduced}. But these methods would reduce
the flexibility of OOMs for dynamic modeling because of their limited
expressive capacity. Recently, a kernel embedding based spectral algorithm
was proposed to cope with continuous data \cite{boots2010hilbert},
which avoids explicit density estimation and learns OOMs in a nonparametric
manner. However, the kernel embedding usually yields a very large
computational complexity, which greatly limits practical applications
of this algorithm to real-world systems.

The purpose of this paper is to address the challenge of spectral
learning of OOMs from nonequilibrium data for analysis of both discrete-
and continuous-valued systems. We first provide a modified spectral
method for discrete-valued stochastic systems which allows us to consistently
estimate the equilibrium dynamics from nonequilibrium data, and then
extend this method to continuous observations in a binless manner.
In comparison with the existing learning methods for continuous OOMs,
the proposed binless spectral method does not rely on any density
estimator, and can achieve consistent estimation with linear computational
complexity in data size even if the assumption of identically distributed
observations does not hold. Moreover, some numerical experiments are
provided to demonstrate the capability of the proposed methods.

\section{Preliminaries}

\subsection{Notation}

In this paper, we use $\mathbb{P}$ to denote probability distribution
for discrete random variables and probability density for continuous
random variables. The indicator function of event $e$ is denoted
by $1_{e}$ and the Dirac delta function centered at $x$ is denoted
by $\delta_{x}\left(\cdot\right)$. For a given process $\{a_{t}\}$,
we write the subsequence $(a_{k},a_{k+1},\ldots,a_{k^{\prime}})$
as $a_{k:k^{\prime}}$, and $\mathbb{E}_{\infty}[a_{t}]\triangleq\lim_{t\to\infty}\mathbb{E}[a_{t}]$
means the equilibrium expected value of $a_{t}$ if the limit exists.
In addition, the convergence in probability is denoted by $\stackrel{p}{\to}$.

\subsection{Observable operator models\label{subsec:Observable-operator-models}}

An $m$-dimensional observable operator model (OOM) with observation
space $\mathcal{O}$ can be represented by a tuple $\mathcal{M}=(\boldsymbol{\omega},\{\boldsymbol{\Xi}(x)\}_{x\in\mathcal{O}},\boldsymbol{\sigma})$,
which consists of an initial state vector $\boldsymbol{\omega}\in\mathbb{R}^{1\times m}$,
an evaluation vector $\boldsymbol{\sigma}\in\mathbb{R}^{m\times1}$
and an observable operator matrix $\boldsymbol{\Xi}(x)\in\mathbb{R}^{m\times m}$
associated to each element $x\in\mathcal{O}$. $\mathcal{M}$ defines
a stochastic process $\{x_{t}\}$ in $\mathcal{O}$ as
\begin{equation}
\mathbb{P}\left(x_{1:t}|\mathcal{M}\right)=\boldsymbol{\omega}\boldsymbol{\Xi}(x_{1:t})\boldsymbol{\sigma}\label{eq:OOM-prob}
\end{equation}
under the condition that $\boldsymbol{\omega}\boldsymbol{\Xi}(x_{1:t})\boldsymbol{\sigma}\ge0$,
$\boldsymbol{\omega}\boldsymbol{\Xi}(\mathcal{O})\boldsymbol{\sigma}=1$
and $\boldsymbol{\omega}\boldsymbol{\Xi}(x_{1:t})\boldsymbol{\sigma}=\boldsymbol{\omega}\boldsymbol{\Xi}(x_{1:t})\boldsymbol{\Xi}(\mathcal{O})\boldsymbol{\sigma}$
hold for all $t$ and $x_{1:t}\in\mathcal{O}^{t}$ \cite{thon2015links},
where $\boldsymbol{\Xi}(x_{1:t})\triangleq\boldsymbol{\Xi}(x_{1})\ldots\boldsymbol{\Xi}(x_{t})$
and $\boldsymbol{\Xi}(\mathcal{A})\triangleq\int_{\mathcal{A}}\mathrm{d}x\ \boldsymbol{\Xi}\left(x\right)$.
Two OOMs $\mathcal{M}$ and $\mathcal{M}^{\prime}$ are said to be
equivalent if $\mathbb{P}\left(x_{1:t}|\mathcal{M}\right)\equiv\mathbb{P}\left(x_{1:t}|\mathcal{M}^{\prime}\right)$.

\section{Spectral learning of OOMs}

\subsection{Algorithm}

\begin{figure}
\begin{centering}
\includegraphics[width=0.4\columnwidth,height=0.13\columnwidth]{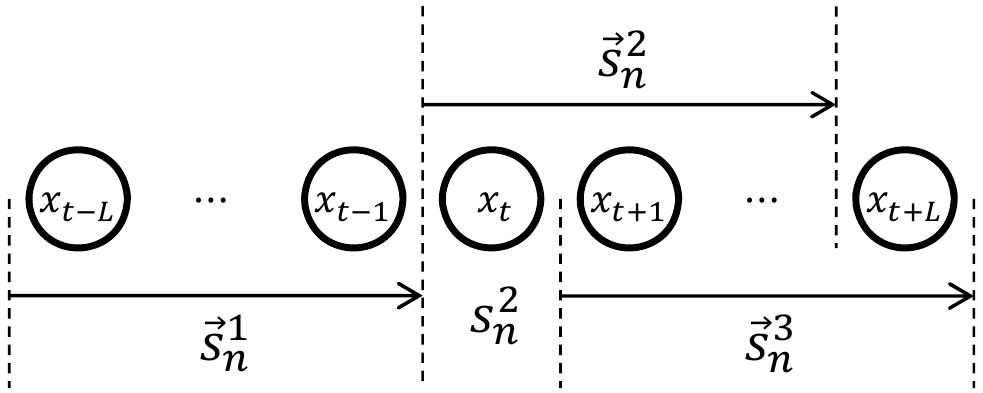}
\par\end{centering}
\caption{\label{fig:Illustration-of-variables}Illustration of variables $\vec{s}_{n}^{\,1}$,
$s_{n}^{2}$, $\vec{s}_{n}^{\,3}$ and $\vec{s}_{n}^{\,2}$ used in
Eqs. (\ref{eq:mean-phi})-(\ref{eq:mean-C13}) with $(\vec{s}_{n}^{\,1},s_{n}^{2},\vec{s}_{n}^{\,3})=x_{t-L:t+L}$.}
\end{figure}

Here and hereafter, we only consider the case that the observation
space $\mathcal{O}$ is a finite set. (Learning with continuous observations
will be discussed in Section \ref{sec:Binless-learning}.) A large
number of largely similar spectral methods have been developed, and
the generic learning procedure of these methods is summarized in Algorithm
\ref{alg:General-procedure} by omitting details of algorithm implementation
and parameter choice \cite{rosencrantz2004learning,hsu2009spectral,boots2012spectral}.
 For convenience of description and analysis, we specify in this
paper the formula for calculating $\hat{\bar{\boldsymbol{\phi}}}_{1}$,
$\hat{\bar{\boldsymbol{\phi}}}_{2}$, $\hat{\mathbf{C}}_{1,2}$ and
$\hat{\mathbf{C}}_{1,3}\left(x\right)$ in Line \ref{lin:statistics}
of Algorithm \ref{alg:General-procedure} as follows:
\begin{gather}
\hat{\bar{\boldsymbol{\phi}}}_{1}=\frac{1}{N}\sum_{n=1}^{N}\boldsymbol{\phi}_{1}(\vec{s}_{n}^{\,1}),\quad\hat{\bar{\boldsymbol{\phi}}}_{2}=\frac{1}{N}\sum_{n=1}^{N}\boldsymbol{\phi}_{2}(\vec{s}_{n}^{\,2})\label{eq:mean-phi}\\
\hat{\mathbf{C}}_{1,2}=\frac{1}{N}\sum_{n=1}^{N}\boldsymbol{\phi}_{1}(\vec{s}_{n}^{\,1})\boldsymbol{\phi}_{2}(\vec{s}_{n}^{\,2})^{\top}\label{eq:mean-C12}\\
\hat{\mathbf{C}}_{1,3}\left(x\right)=\frac{1}{N}\sum_{n=1}^{N}1_{s_{n}^{2}=x}\boldsymbol{\phi}_{1}(\vec{s}_{n}^{\,1})\boldsymbol{\phi}_{2}(\vec{s}_{n}^{\,3})^{\top},\quad\forall x\in\mathcal{O}\label{eq:mean-C13}
\end{gather}
Here $\{(\vec{s}_{n}^{\,1},s_{n}^{2},\vec{s}_{n}^{\,3})\}_{n=1}^{N}$
is the collection of all subsequences of length $\left(2L+1\right)$
appearing in observation data ($N=T-2L$ for a single observation
trajectory of length $T$). If an observation subsequence $x_{t-L:t+L}$
is denoted by $(\vec{s}_{n}^{\,1},s_{n}^{2},\vec{s}_{n}^{\,3})$ with
some $n$, then $\vec{s}_{n}^{\,1}=x_{t-L:t-1}$ and $\vec{s}_{n}^{\,3}=x_{t+1:t+L}$
represents the prefix and suffix of $x_{t-L:t+L}$ of length $L$,
$s_{n}^{2}=x_{t}$ is the intermediate observation value, and $\vec{s}_{n}^{\,2}=x_{t:t+L-1}$
is an ``intermediate part'' of the subsequence of length $L$ starting
from time $t$ (see Fig.~\ref{fig:Illustration-of-variables} for
a graphical illustration).

\begin{algorithm}
\caption{General procedure for spectral learning of OOMs\label{alg:General-procedure}}

\begin{algorithmic}[1]

\INPUT Observation trajectories generated by a stochastic process
$\{x_{t}\}$ in $\mathcal{O}$

\OUTPUT $\hat{\mathcal{M}}=(\hat{\boldsymbol{\omega}},\{\hat{\boldsymbol{\Xi}}(x)\}_{x\in\mathcal{O}},\hat{\boldsymbol{\sigma}})$

\PARAMETER $m$: dimension of the OOM. $D_{1},D_{2}$: numbers of
feature functions. $L$: order of feature functions.

\State Construct feature functions $\boldsymbol{\phi}_{1}=(\varphi_{1,1},\ldots,\varphi_{1,D_{1}})^{\top}$
and $\boldsymbol{\phi}_{2}=(\varphi_{2,1},\ldots,\varphi_{2,D_{2}})^{\top}$,
where each $\varphi_{i,j}$ is a mapping from $\mathcal{O}^{L}$ to
$\mathbb{R}$ and $D_{1},D_{2}\ge m$.

\State\label{lin:statistics} Approximate
\begin{gather}
\bar{\boldsymbol{\phi}}_{1}\triangleq\mathbb{E}\left[\boldsymbol{\phi}_{1}(x_{t-L:t-1})\right],\quad\bar{\boldsymbol{\phi}}_{2}\triangleq\mathbb{E}\left[\boldsymbol{\phi}_{2}(x_{t:t+L-1})\right]\\
\mathbf{C}_{1,2}\triangleq\mathbb{E}\left[\boldsymbol{\phi}_{1}(x_{t-L:t-1})\boldsymbol{\phi}_{2}(x_{t:t+L-1})^{\top}\right]\\
\mathbf{C}_{1,3}\left(x\right)\triangleq\mathbb{E}\left[1_{x_{t}=x}\cdot\boldsymbol{\phi}_{1}(x_{t-L:t-1})\boldsymbol{\phi}_{2}(x_{t+1:t+L})^{\top}\right],\quad\forall x\in\mathcal{O}
\end{gather}
by their empirical means $\hat{\bar{\boldsymbol{\phi}}}_{1}$, $\hat{\bar{\boldsymbol{\phi}}}_{2}$,
$\hat{\mathbf{C}}_{1,2}$ and $\hat{\mathbf{C}}_{1,3}\left(x\right)$
over observation data.

\State Compute $\mathbf{F}_{1}=\mathbf{U}\boldsymbol{\Sigma}^{-1}\in\mathbb{R}^{D_{1}\times m}$
and $\mathbf{F}_{2}=\mathbf{V}\in\mathbb{R}^{D_{2}\times m}$ from
the truncated singular value decomposition $\hat{\mathbf{C}}_{1,2}\approx\mathbf{U}\boldsymbol{\Sigma}\mathbf{V}^{\top}$,
where $\boldsymbol{\Sigma}\in\mathbb{R}^{m\times m}$ is a diagonal
matrix contains the top $m$ singular values of $\hat{\mathbf{C}}_{1,2}$,
and $\mathbf{U}$ and $\mathbf{V}$ consist of the corresponding $m$
left and right singular vectors of $\hat{\mathbf{C}}_{1,2}$.

\State Compute

\begin{eqnarray}
\hat{\boldsymbol{\sigma}} & = & \mathbf{F}_{1}^{\top}\hat{\bar{\boldsymbol{\phi}}}_{1}\label{eq:sigma-hat}\\
\hat{\boldsymbol{\Xi}}(x) & = & \mathbf{F}_{1}^{\top}\hat{\mathbf{C}}_{1,3}(x)\mathbf{F}_{2},\quad\forall x\in\mathcal{O}\label{eq:Xi-hat}\\
\hat{\boldsymbol{\omega}} & = & \hat{\bar{\boldsymbol{\phi}}}_{2}^{\top}\mathbf{F}_{2}
\end{eqnarray}


\end{algorithmic}
\end{algorithm}

Algorithm \ref{alg:General-procedure} is much more efficient than
the commonly used likelihood-based learning algorithms and does not
suffer from local optima issues. In addition, and more importantly,
this algorithm can be shown to be consistent if $(\vec{s}_{n}^{\,1},s_{n}^{2},\vec{s}_{n}^{\,3})$
are (i) independently sampled from $\mathcal{M}$ or (ii) obtained
from a finite number of trajectories which have fully mixed so that
all observation triples are identically distributed (see, e.g., \cite{siddiq2010reduced,Jaeger2012Discrete,thon2015links}
for related works). However, the asymptotic correctness of OOMs learned
from short trajectories starting from nonequilibrium states has not
been formally determined.

\subsection{Theoretical analysis}

We now analyze statistical properties of the spectral algorithm without
the assumption of identically distributed observations. Before stating
our main result, some assumptions on observation data are listed as
follows:
\begin{assumption}
\label{assu:infinity}The observation data consists of $I$ independent
trajectories of length $T$ produced by a stochastic process $\{x_{t}\}$,
and the data size tends to infinity with (i) $I\to\infty$ and $T=T_{0}$
or (ii) $T\to\infty$ and $I=I_{0}$.
\end{assumption}

\begin{assumption}
\label{assu:stationary}$\{x_{t}\}$ is driven by an $m$-dimensional
OOM $\mathcal{M}=(\boldsymbol{\omega},\{\boldsymbol{\Xi}(x)\}_{x\in\mathcal{O}},\boldsymbol{\sigma})$,
and

\begin{equation}
\frac{1}{T^{\prime}}\sum_{t=1}^{T^{\prime}}f_{t}\stackrel{p}{\to}\mathbb{E}_{\infty}\left[f\left(x_{t:t+l-1}\right)\right]=\mathbb{E}_{\infty}\left[f\left(x_{t:t+l-1}\right)|x_{1:k}\right]
\end{equation}
as $T^{\prime}\to\infty$ for all $k$, $l$, $x_{1:k}$ and $f:\mathcal{O}^{l}\mapsto\mathbb{R}$.
\end{assumption}

\begin{assumption}
\label{assu:invertible}The rank of the limit of $\hat{\mathbf{C}}_{1,2}$
is not less than $m$.
\end{assumption}
Notice that Assumption \ref{assu:stationary} only states the asymptotic
stationarity of $\{x_{t}\}$ and marginal distributions of observation
triples are possibly time dependent if $\boldsymbol{\omega}\neq\boldsymbol{\omega}\boldsymbol{\Xi}\left(\mathcal{O}\right)$.
Assumption \ref{assu:invertible} ensures that the limit of $\hat{\mathcal{M}}$
given by Algorithm \ref{alg:General-procedure} is well defined, which
generally holds for minimal OOMs (see \cite{thon2015links}).

Based on the above assumptions, we have the following theorem concerning
the statistical consistency of the OOM learning algorithm (see Appendix
\ref{subsec:Proof-of-Theorem-nonequilibrium} for proof):
\begin{thm}
\label{thm:nonequilibrium}Under Assumptions \ref{assu:infinity}-\ref{assu:invertible},
there exists an OOM $\mathcal{M}^{\prime}=(\boldsymbol{\omega}^{\prime},\{\boldsymbol{\Xi}^{\prime}(x)\}_{x\in\mathcal{O}},\boldsymbol{\sigma}^{\prime})$
which is equivalent to $\hat{\mathcal{M}}$ and satisfies
\begin{equation}
\boldsymbol{\sigma}^{\prime}\stackrel{p}{\to}\boldsymbol{\sigma},\quad\boldsymbol{\Xi}^{\prime}(x)\stackrel{p}{\to}\boldsymbol{\Xi}(x),\:\forall x\in\mathcal{O}
\end{equation}
\end{thm}
This theorem is central in this paper, which implies that the spectral
learning algorithm can achieve consistent estimation of all parameters
of OOMs except initial state vectors even for nonequilibrium data.
($\hat{\boldsymbol{\omega}}\stackrel{p}{\to}\boldsymbol{\omega}^{\prime}$
does not hold in most cases except when $\{x_{t}\}$ is stationary.).
It can be further generalized according to requirements in more complicated
situations where, for example, observation trajectories are generated
with multiple different initial conditions (see Appendix \ref{subsec:Asymptotic-correctness-different-initial}).

\section{Spectral learning of equilibrium OOMs}

In this section, applications of spectral learning to the problem
of recovering equilibrium properties of dynamic systems from nonequilibrium
data will be highlighted, which is an important problem in practice
especially for thermodynamic and kinetic analysis in computational
physics and chemistry.

\subsection{Learning from discrete data\label{sec:Nonequilibrium-learning}}

According to the definition of OOMs, the equilibrium dynamics of an
OOM $\mathcal{M}=(\boldsymbol{\omega},\{\boldsymbol{\Xi}(x)\}_{x\in\mathcal{O}},\boldsymbol{\sigma})$
can be described by an equilibrium OOM $\mathcal{M}_{\mathrm{eq}}=(\boldsymbol{\omega}_{\mathrm{eq}},\{\boldsymbol{\Xi}(x)\}_{x\in\mathcal{O}},\boldsymbol{\sigma})$
as
\begin{equation}
\lim_{t\to\infty}\mathbb{P}\left(x_{t+1:t+k}=z_{1:k}|\mathcal{M}\right)=\mathbb{P}\left(x_{1:t}=z_{1:k}|\mathcal{M}_{\mathrm{eq}}\right)\label{eq:equilibrium-dynamics}
\end{equation}
if the equilibrium state vector
\begin{equation}
\boldsymbol{\omega}_{\mathrm{eq}}=\lim_{t\to\infty}\boldsymbol{\omega}\boldsymbol{\Xi}(\mathcal{O})^{t}\label{eq:omega-eq}
\end{equation}
exists. From (\ref{eq:equilibrium-dynamics}) and (\ref{eq:omega-eq}),
we have
\begin{equation}
\left\{ \begin{array}{l}
\boldsymbol{\omega}_{\mathrm{eq}}\boldsymbol{\Xi}(\mathcal{O})=\lim_{t\to\infty}\boldsymbol{\omega}_{\mathrm{eq}}\boldsymbol{\Xi}(\mathcal{O})^{t+1}=\boldsymbol{\omega}_{\mathrm{eq}}\\
\boldsymbol{\omega}_{\mathrm{eq}}\boldsymbol{\sigma}=\lim_{t\to\infty}\sum_{x\in\mathcal{O}}\mathbb{P}\left(x_{t+1}=x\right)=1
\end{array}\right.\label{eq:omega-eq-bar-constraint}
\end{equation}
The above equilibrium constraint of OOMs motivates the following algorithm
for learning equilibrium OOMs: \emph{Perform Algorithm \ref{alg:General-procedure}
to get $\hat{\boldsymbol{\Xi}}\left(x\right)$ and $\hat{\boldsymbol{\sigma}}$
and calculate $\hat{\boldsymbol{\omega}}_{\mathrm{eq}}$ by a quadratic
programming problem 
\begin{equation}
\hat{\boldsymbol{\omega}}_{\mathrm{eq}}=\arg\min_{\mathbf{w}\in\{\mathbf{w}|\mathbf{w}\hat{\boldsymbol{\sigma}}=1\}}\left\Vert \mathbf{w}\hat{\boldsymbol{\Xi}}(\mathcal{O})-\mathbf{w}\right\Vert ^{2}\label{eq:nonequilibrium-omega-opt}
\end{equation}
}(See Appendix \ref{subsec:Proof-of-Corollary-nonequilibrium-omega}
for a closed-form expression of the solution to (\ref{eq:nonequilibrium-omega-opt}).)

The existence and uniqueness of $\boldsymbol{\omega}_{\mathrm{eq}}$
are shown in Appendix \ref{subsec:Proof-of-Corollary-nonequilibrium-omega},
which yield the following theorem:
\begin{thm}
\label{thm:nonequilibrium-omega}Under Assumptions \ref{assu:infinity}-\ref{assu:invertible},
the estimated equilibrium OOM $\hat{\mathcal{M}}_{\mathrm{eq}}=(\hat{\boldsymbol{\omega}}_{\mathrm{eq}},\{\hat{\boldsymbol{\Xi}}(x)\}_{x\in\mathcal{O}},\hat{\boldsymbol{\sigma}})$
provided by Algorithm \ref{alg:General-procedure} and Eq.~(\ref{eq:nonequilibrium-omega-opt})
satisfies
\begin{equation}
\mathbb{P}\left(x_{1:l}=z_{1:l}|\hat{\mathcal{M}}_{\mathrm{eq}}\right)\stackrel{p}{\to}\lim_{t\to\infty}\mathbb{P}\left(x_{t+1:t+l}=z_{1:l}\right)\label{eq:oom-convergence}
\end{equation}
for all $l$ and $z_{1:l}$.
\end{thm}
\begin{rem}
$\hat{\boldsymbol{\omega}}_{\mathrm{eq}}$ can also be computed as
an eigenvector of $\hat{\boldsymbol{\Xi}}(\mathcal{O})$. But the
eigenvalue problem possibly yields numerical instability and complex
values because of statistical noise, unless some specific feature
functions $\boldsymbol{\phi}_{1},\boldsymbol{\phi}_{2}$ are selected
so that $\hat{\boldsymbol{\omega}}_{\mathrm{eq}}\hat{\boldsymbol{\Xi}}(\mathcal{O})=\hat{\boldsymbol{\omega}}_{\mathrm{eq}}$
can be exactly solved in the real field \cite{jaeger2005efficient}.
\end{rem}

\subsection{Learning from continuous data\label{sec:Binless-learning}}

A straightforward way to extend spectral algorithms to handle continuous
data is based on the coarse-graining of the observation space. Suppose
that $\{x_{t}\}$ is a stochastic process in a continuous observation
space $\mathcal{O}\subset\mathbb{R}^{d}$, and $\mathcal{O}$ is partitioned
into $J$ discrete bins $\mathcal{B}_{1},\ldots,\mathcal{B}_{J}$.
Then we can utilize the algorithm in Section \ref{sec:Nonequilibrium-learning}
to approximate the equilibrium transition dynamics between bins as
\begin{equation}
\lim_{t\to\infty}\mathbb{P}\left(x_{t+1}\in\mathcal{B}_{j_{1}},\ldots,x_{t+l}\in\mathcal{B}_{j_{l}}\right)\approx\hat{\boldsymbol{\omega}}_{\mathrm{eq}}\hat{\boldsymbol{\Xi}}\left(\mathcal{B}_{j_{1}}\right)\ldots\hat{\boldsymbol{\Xi}}\left(\mathcal{B}_{j_{l}}\right)\hat{\boldsymbol{\sigma}}
\end{equation}
and obtain a \emph{binned} OOM $\hat{\mathcal{M}}_{\mathrm{eq}}=(\hat{\boldsymbol{\omega}}_{\mathrm{eq}},\{\hat{\boldsymbol{\Xi}}(x)\}_{x\in\mathcal{O}},\hat{\boldsymbol{\sigma}})$
for the continuous dynamics of $\{x_{t}\}$ with
\begin{equation}
\hat{\boldsymbol{\Xi}}(x)=\frac{\hat{\boldsymbol{\Xi}}(\mathcal{B}\left(x\right))}{\mathrm{vol}(\mathcal{B}\left(x\right))}
\end{equation}
by assuming the observable operator matrices are piecewise constant
on bins, where $\mathcal{B}\left(x\right)$ denotes the bin containing
$x$ and $\mathrm{vol}(\mathcal{B})$ is the volume of $\mathcal{B}$.
Conventional wisdom dictates that the number of bins is a key parameter
for the coarse-graining strategy and should be carefully chosen for
the balance of statistical noise and discretization error. However,
we will show in what follows that it is justifiable to increase the
number of bins to infinity.

Let us consider the limit case where $J\to\infty$ and bins are infinitesimal
with $\max_{j}\mathrm{vol}(\mathcal{B}_{j})\to0$. In this case,
\begin{equation}
\hat{\boldsymbol{\Xi}}(x)=\lim_{\mathrm{vol}(\mathcal{B}\left(x\right))\to0}\frac{\hat{\boldsymbol{\Xi}}(\mathcal{B}\left(x\right))}{\mathrm{vol}(\mathcal{B}\left(x\right))}=\left\{ \begin{array}{ll}
\hat{\mathbf{W}}_{s_{n}^{2}}\delta_{s_{n}^{2}}\left(x\right), & x=s_{n}^{2}\\
0, & \mathrm{otherwise}
\end{array}\right.
\end{equation}
where
\begin{equation}
\hat{\mathbf{W}}_{s_{n}^{2}}=\frac{1}{N}\mathbf{F}_{1}^{\top}\boldsymbol{\phi}_{1}(\vec{s}_{n}^{\,1})\boldsymbol{\phi}_{2}(\vec{s}_{n}^{\,3})^{\top}\mathbf{F}_{2}\label{eq:W}
\end{equation}
according to (\ref{eq:Xi-hat}) in Algorithm \ref{alg:General-procedure}.
Then $\hat{\mathcal{M}}_{\mathrm{eq}}$ becomes a \emph{binless} OOM
over sample points $\mathcal{X}=\{s_{n}^{2}\}_{n=1}^{N}$ and can
be estimated from data by Algorithm \ref{alg:binless-procedure},
where the feature functions can be selected as indicator functions,
radial basis functions or other commonly used activation functions
for single-layer neural networks in order to digest adequate dynamic
information from observation data.

\begin{algorithm}
\caption{Procedure for learning binless equilibrium OOMs\label{alg:binless-procedure}}

\begin{algorithmic}[1]

\INPUT Observation trajectories generated by a stochastic process
$\{x_{t}\}$ in $\mathcal{O}\subset\mathbb{R}^{d}$

\OUTPUT Binless OOM $\hat{\mathcal{M}}=(\hat{\boldsymbol{\omega}},\{\hat{\boldsymbol{\Xi}}(x)\}_{x\in\mathcal{O}},\hat{\boldsymbol{\sigma}})$

\State Construct feature functions $\boldsymbol{\phi}_{1}:\mathbb{R}^{Ld}\mapsto\mathbb{R}^{D_{1}}$
and $\boldsymbol{\phi}_{2}:\mathbb{R}^{Ld}\mapsto\mathbb{R}^{D_{2}}$
with $D_{1},D_{2}\ge m$.

\State Calculate $\hat{\bar{\boldsymbol{\phi}}}_{1},\hat{\bar{\boldsymbol{\phi}}}_{2},\hat{\mathbf{C}}_{1,2}$
by (\ref{eq:mean-phi}) and (\ref{eq:mean-C12}).

\State Compute $\mathbf{F}_{1}=\mathbf{U}\boldsymbol{\Sigma}^{-1}\in\mathbb{R}^{D_{1}\times m}$
and $\mathbf{F}_{2}=\mathbf{V}\in\mathbb{R}^{D_{2}\times m}$ from
the truncated singular value decomposition $\hat{\mathbf{C}}_{1,2}\approx\mathbf{U}\boldsymbol{\Sigma}\mathbf{V}^{\top}$.

\State Compute $\hat{\boldsymbol{\sigma}},\hat{\boldsymbol{\omega}}$
and $\hat{\boldsymbol{\Xi}}(x)=\sum_{z\in\mathcal{X}}\hat{\mathbf{W}}_{z}\delta_{z}\left(x\right)$
by (\ref{eq:sigma-hat}), (\ref{eq:nonequilibrium-omega-opt}) and
(\ref{eq:W}), where $\hat{\boldsymbol{\Xi}}(\mathcal{O})=\int_{\mathcal{O}}\mathrm{d}x\ \hat{\boldsymbol{\Xi}}(x)=\sum_{z\in\mathcal{X}}\hat{\mathbf{W}}_{z}$.

\end{algorithmic}
\end{algorithm}

The binless algorithm presented here can be efficiently implemented
in a linear computational complexity $O(N)$, and is applicable to
more general cases where observations are strings, graphs or other
structured variables. Unlike the other spectral algorithms for continuous
data, it does not require that the observed dynamics coincides with
some parametric model defined by feature functions. Lastly but most
importantly, as stated in the following theorem, this algorithm can
be used to consistently extract static and kinetic properties of a
dynamic system in equilibrium from nonequilibrium data (see Appendix
\ref{subsec:Proof-of-Corollary-nonequilibrium-omega} for proof):
\begin{thm}
\label{thm:bloom}Provided that the observation space $\mathcal{O}$
is a closed set in $\mathbb{R}^{d}$, feature functions $\boldsymbol{\phi}_{1},\boldsymbol{\phi}_{2}$
are bounded on $\mathcal{O}^{L}$, and Assumptions \ref{assu:infinity}-\ref{assu:invertible}
hold, the binless OOM given by Algorithm \ref{alg:binless-procedure}
satisfies
\begin{equation}
\mathbb{E}\left[g\left(x_{1:r}\right)|\hat{\mathcal{M}}_{\mathrm{eq}}\right]\stackrel{p}{\to}\mathbb{E}_{\infty}\left[g\left(x_{t+1:t+r}\right)\right]
\end{equation}
with
\begin{equation}
\mathbb{E}\left[g\left(x_{1:r}\right)|\hat{\mathcal{M}}_{\mathrm{eq}}\right]=\sum_{x_{1:r}\in\mathcal{X}^{r}}g\left(x_{1:r}\right)\hat{\boldsymbol{\omega}}\hat{\mathbf{W}}_{z_{1}}\ldots\hat{\mathbf{W}}_{z_{r}}\hat{\boldsymbol{\sigma}}\label{eq:binless-expectation}
\end{equation}

\begin{enumerate}
\item[(i)] for all continuous functions $g:\mathcal{O}^{r}\mapsto\mathbb{R}$.
\item[(ii)] for all bounded and Borel measurable functions $g:\mathcal{O}^{r}\mapsto\mathbb{R}$,
if there exist positive constants $\bar{\xi}$ and $\underline{\xi}$
so that $\left\Vert \boldsymbol{\Xi}\left(x\right)\right\Vert \le\bar{\xi}$
and $\lim_{t\to\infty}\mathbb{P}\left(x_{t+1:t+r}=z_{1:r}\right)\ge\underline{\xi}$
for all $x\in\mathcal{O}$ and $z_{1:r}\in\mathcal{O}^{r}$.
\end{enumerate}
\end{thm}

\subsection{Comparison with related methods}

It is worth pointing out that the spectral learning investigated in
this section is an ideal tool for analysis of dynamic properties of
stochastic processes, because the related quantities, such as stationary
distributions, principle components and time-lagged correlations,
can be easily computed from parameters of discrete OOMs or binless
OOMs. For many popular nonlinear dynamic models, including Gaussian
process state-space models \cite{turner2010state} and recurrent neural
networks \cite{hochreiter1997long}, the computation of such quantities
is intractable or time-consuming.

The major disadvantage of spectral learning is that the estimated
OOMs are usually only ``approximately valid'' and possibly assign
``negative probabilities'' to some observation sequences. So it
is difficult to apply spectral methods to prediction, filtering and
smoothing of signals where the Bayesian inference is involved.

\section{Applications\label{sec:Applications}}

In this section, we evaluate our algorithms on two diffusion processes
and the molecular dynamics of alanine dipeptide, and compare them
to several alternatives. The detailed settings of simulations and
algorithms are provided in Appendix \ref{sec:Settings-in-applications}.

\paragraph{Brownian dynamics}

\begin{figure}
\begin{centering}
\includegraphics[height=0.5\textwidth]{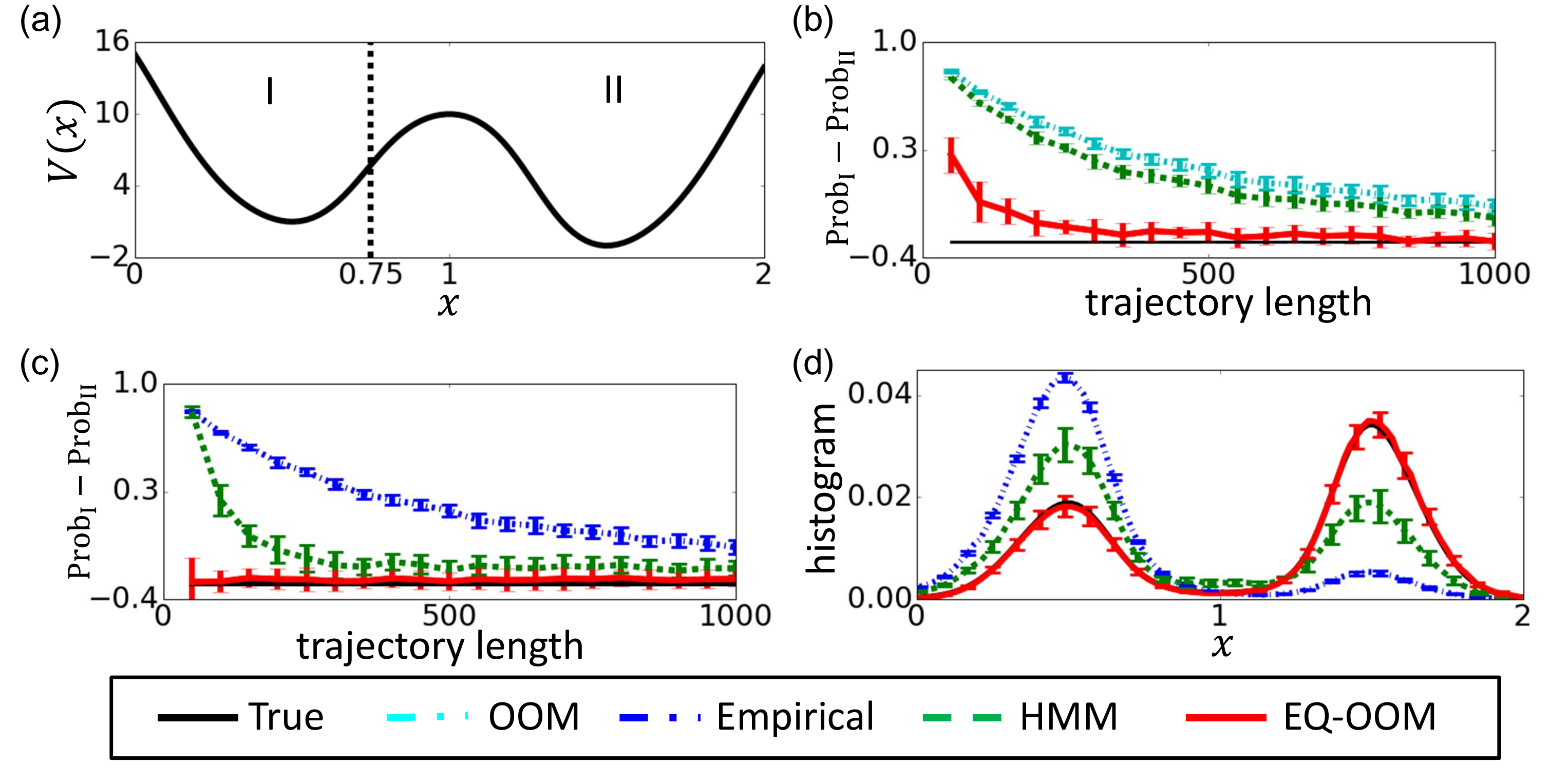}
\par\end{centering}
\caption{Comparison of modeling methods for a one-dimensional diffusion process.
(a) Potential function. (b) Estimates of the difference between equilibrium
probabilities of I and II given by the traditional OOM, HMM and the
equilibrium OOM (EQ-OOM) obtained from the proposed algorithm with
$\mathcal{O}=\{\mathrm{I},\mathrm{II}\}$. (c) Estimates of the probability
difference given by the empirical estimator, HMM and the proposed
binless OOM with $\mathcal{O}=[0,2]$. (d) Stationary histograms of
$\{x_{t}\}$ with $100$ uniform bins estimated from trajectories
with length $50$. The length of each trajectory is $T=50\sim1000$
and the number of trajectories is $[10^{5}/T]$. Error bars are standard
deviations over $30$ independent experiments.\label{fig:diffusion-1d}}
\end{figure}

Let us consider a one-dimensional diffusion process driven by the
Brownian dynamics
\begin{equation}
\mathrm{d}x_{t}=-\nabla V(x_{t})\mathrm{d}t+\sqrt{2\beta^{-1}}\mathrm{d}W_{t}\label{eq:brownian-dynamics}
\end{equation}
with observations generated by
\[
y_{t}=\left\{ \begin{array}{ll}
1, & x_{t}\in\mathrm{I}\\
0, & x_{t}\in\mathrm{II}
\end{array}\right.
\]
The potential function $V\left(x\right)$ is shown in Fig.~\ref{fig:diffusion-1d}(a),
which contains two potential wells $\mathrm{I},\mathrm{II}$. In this
example, all simulations are performed by starting from a uniform
distribution on $[0,0.2]$, which implies that simulations are highly
nonequilibrium and it is difficult to accurately estimate the equilibrium
probabilities $\mathrm{Prob_{I}}=\mathbb{E}_{\infty}\left[1_{x_{t}\in\mathrm{I}}\right]=\mathbb{E}_{\infty}\left[y_{t}\right]$
and $\mathrm{Prob_{II}}=\mathbb{E}_{\infty}\left[1_{x_{t}\in\mathrm{II}}\right]=1-\mathbb{E}_{\infty}\left[y_{t}\right]$
of the two potential wells from the simulation data. We first utilize
the traditional spectral learning without enforcing equilibrium, expectation–maximization
based HMM learning and the proposed discrete spectral algorithm to
estimate $\mathrm{Prob_{I}}$ and $\mathrm{Prob_{II}}$ based on $\{y_{t}\}$,
and the estimation results with different simulation lengths are summarized
in Fig.~\ref{fig:diffusion-1d}(b). It can be seen that, in contrast
to with the other methods, the spectral algorithm for equilibrium
OOMs effectively reduce the statistical bias in the nonequilibrium
data, and achieves statistically correct estimation at $T=300$.

Figs.~\ref{fig:diffusion-1d}(c) and \ref{fig:diffusion-1d}(d) plot
estimates of stationary distribution of $\{x_{t}\}$ obtained from
$\{x_{t}\}$ directly, where the empirical estimator calculates statistics
through averaging over all observations. In this case, the proposed
binless OOM significantly outperform the other methods, and its estimates
are very close to true values even for extremely small short trajectories.

\begin{figure}
\begin{centering}
\includegraphics[width=1\textwidth]{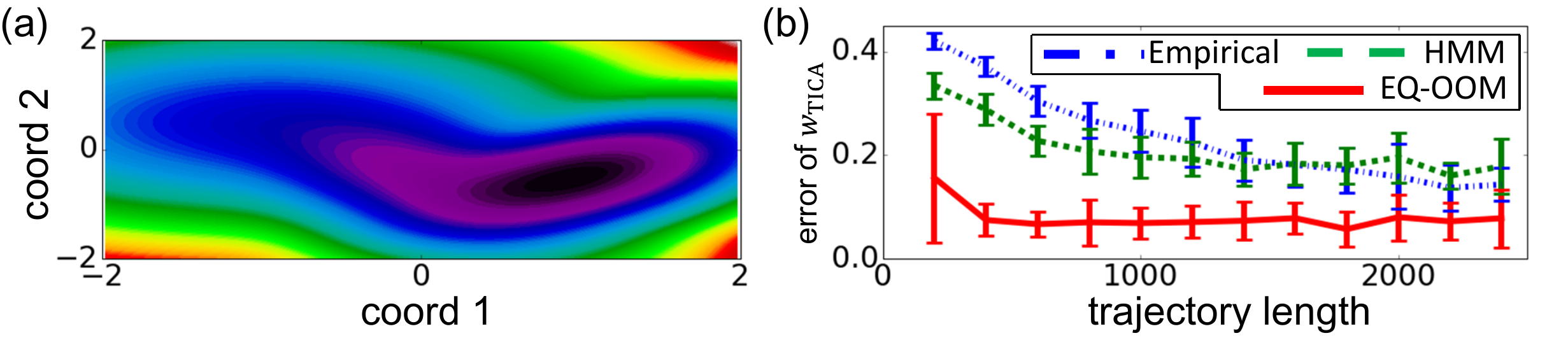}
\par\end{centering}
\caption{Comparison of modeling methods for a two-dimensional diffusion process.
(a) Potential function. (b) Estimation error of $w_{\mathrm{TICA}}\in\mathbb{R}^{2}$
of the first TIC with lag time $100$. Length of each trajectory is
$T=200\sim2500$ and the number of trajectories is $[10^{5}/T]$.
Error bars are standard deviations over $30$ independent experiments.\label{fig:diffusion-2d}}
\end{figure}

Fig.~\ref{fig:diffusion-2d} provides an example of a two-dimensional
diffusion process. The dynamics of this process can also be represented
in the form of (\ref{eq:brownian-dynamics}) and the potential function
is shown in Fig.~\ref{fig:diffusion-2d}(a). The goal of this example
is to estimate the first time-structure based independent component
$w_{\mathrm{TICA}}$ \cite{perez2013identification} of this process
from simulation data. Here $w_{\mathrm{TICA}}$ is a kinetic quantity
of the process and is the solution to the generalized eigenvalue problem
\[
C_{\tau}w=\lambda C_{0}w
\]
with the largest eigenvalue, where $C_{0}$ is the covariance matrix
of $\{x_{t}\}$ in equilibrium and $C_{\tau}=\left(\mathbb{E}_{\infty}\left[x_{t}x_{t+\tau}^{\top}\right]-\mathbb{E}_{\infty}\left[x_{t}\right]\mathbb{E}_{\infty}\left[x_{t}^{\top}\right]\right)$
is the equilibrium time-lagged covariance matrix. The simulation data
are also nonequilibrium with all simulations starting from the uniform
distribution on $[-2,0]\times[-2,0]$. Fig.~\ref{fig:diffusion-2d}(b)
displays the estimation errors of $w_{\mathrm{TICA}}$ obtained from
different learning methods, which also demonstrates the superiority
of the binless spectral method.

\paragraph{Alanine dipeptide}

\begin{figure}
\begin{centering}
\includegraphics[width=1\textwidth]{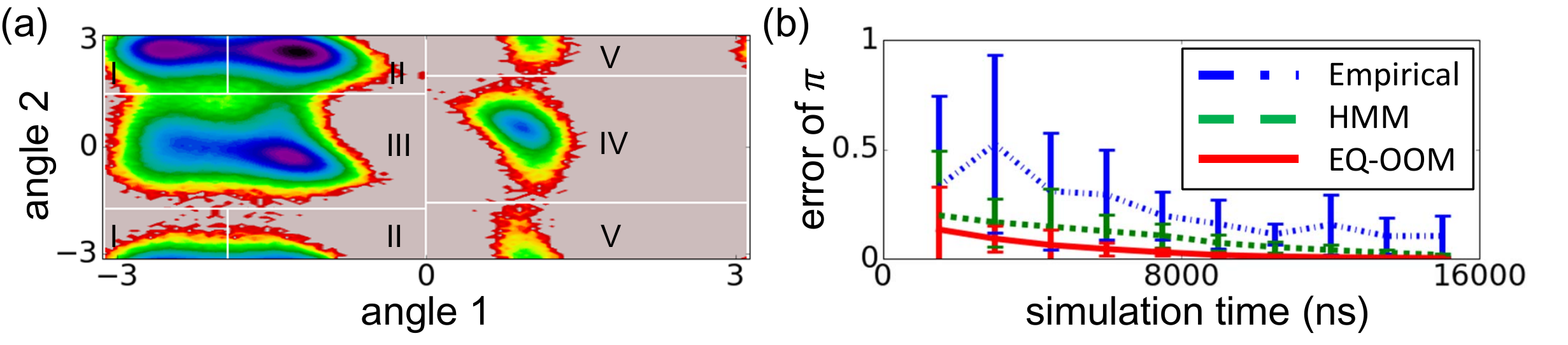}
\par\end{centering}
\caption{Comparison of modeling methods for molecular dynamics of alanine dipeptide.
(a) Reduced free energy. (b) Estimation error of $\boldsymbol{\pi}$,
where the horizontal axis denotes the total simulation time $T\times I$.
Length of each trajectory is $T=10\mathrm{ns}$ and the number of
trajectories is $I=150\sim1500$. Error bars are standard deviations
over $30$ independent experiments.\label{fig:ad}}
\end{figure}

Alanine dipeptide is a small molecule which consists of two alanine
amino acid units, and its configuration can be described by two backbone
dihedral angles. Fig.~\ref{fig:ad}(a) shows the potential profile
of the alanine dipeptide with respect to the two angles, which contains
five metastable states $\{\mathrm{I},\mathrm{II},\mathrm{III},\mathrm{IV},\mathrm{V}\}$.
We perform multiple short molecular dynamics simulations starting
from the metastable state $\mathrm{IV}$, where each simulation length
is $10\mathrm{ns}$, and utilizes different methods to approximate
the stationary distribution $\pi=(\mathrm{Prob_{I}},\mathrm{Prob_{II}},\ldots,\mathrm{Prob_{V}})$
of the five metastable states. As shown in Fig.~\ref{fig:ad}(b),
the proposed binless algorithm yields lower estimation error compared
to each of the alternatives.

\section{Conclusion}

In this paper, we investigated the statistical properties of the general
spectral learning procedure for nonequilibrium data, and developed
novel spectral methods for learning equilibrium dynamics from nonequilibrium
(discrete or continuous) data. The main ideas of the presented methods
are to correct the model parameters by the equilibrium constraint
and to handle continuous observations in a binless manner. Interesting
directions of future research include analysis of approximation error
with finite data size and applications to controlled systems.

\subsubsection*{Acknowledgments}

This work was funded by Deutsche Forschungsgemeinschaft (SFB 1114/A04
to H.W.) and European Research Council (StG pcCell to F.N.). 

\bibliographystyle{ieeetr}
\bibliography{oom_bib_hwu}

\clearpage
\setcounter{page}{1}
\numberwithin{equation}{section}

\appendix

\part*{Supplementary Information}

\section{Proofs\label{sec:Proofs}}

\subsection{Proof of Theorem \ref{thm:nonequilibrium}\label{subsec:Proof-of-Theorem-nonequilibrium}}

For convenience, here we define
\[
\boldsymbol{\omega}^{*}(T)=\frac{1}{T-2L}\boldsymbol{\omega}\sum_{t=1}^{T-2L}\boldsymbol{\Xi}(\mathcal{O})^{t-1}
\]
\begin{equation}
\mathbf{G}_{\sigma}=\sum_{z_{1:L}}\boldsymbol{\phi}_{2}(z_{1:L})\boldsymbol{\sigma}^{\top}\boldsymbol{\Xi}(z_{1:L})^{\top}\label{eq:G-sigma}
\end{equation}

\paragraph{Part (1)}

We first show the theorem in the case of $T=T_{0}$ and $I\to\infty$.

Let
\begin{equation}
\mathbf{G}_{\omega}=\sum_{z_{1:L}}\boldsymbol{\phi}_{1}(z_{1:L})\boldsymbol{\omega}^{*}(T_{0})\boldsymbol{\Xi}(z_{1:L})\label{eq:G-omega-finite}
\end{equation}
Since $I\to\infty$, we have
\begin{eqnarray*}
\hat{\bar{\boldsymbol{\phi}}}_{1} & \stackrel{p}{\to} & \mathbb{E}\left[\hat{\bar{\boldsymbol{\phi}}}_{1}\right]=\mathbf{G}_{\omega}\boldsymbol{\sigma}\\
\hat{\bar{\boldsymbol{\phi}}}_{2}^{\top} & \stackrel{p}{\to} & \mathbb{E}\left[\hat{\bar{\boldsymbol{\phi}}}_{2}^{\top}\right]=\boldsymbol{\omega}^{*}(T_{0})\mathbf{G}_{\sigma}^{\top}\\
\hat{\mathbf{C}}_{1,2} & \stackrel{p}{\to} & \mathbb{E}\left[\hat{\mathbf{C}}_{1,2}\right]=\mathbf{G}_{\omega}\mathbf{G}_{\sigma}^{\top}\\
\hat{\mathbf{C}}_{1,3}\left(x\right) & \stackrel{p}{\to} & \mathbb{E}\left[\hat{\mathbf{C}}_{1,2}\left(x\right)\right]=\mathbf{G}_{\omega}\boldsymbol{\Xi}(x)\mathbf{G}_{\sigma}^{\top}
\end{eqnarray*}

According to Assumption \ref{assu:invertible} and the Eckart-Young-Mirsky
Theorem, we can conclude that
\[
\mathrm{rank}\left(\mathbf{G}_{\omega}\right)=\mathrm{rank}\left(\mathbf{G}_{\sigma}\right)=\mathrm{rank}\left(\hat{\mathbf{C}}_{1,2}\right)=m
\]
and
\[
\hat{\mathbf{C}}_{1,2}^{\mathrm{trun}}=\mathbf{U}\boldsymbol{\Sigma}\mathbf{V}^{\top}\stackrel{p}{\to}\mathbf{G}_{\omega}\mathbf{G}_{\sigma}^{\top}
\]

By using the SVD of $\mathbf{G}_{\omega}\mathbf{G}_{\sigma}^{\top}$
\[
\mathbf{G}_{\omega}\mathbf{G}_{\sigma}^{\top}=\tilde{\mathbf{U}}\tilde{\boldsymbol{\Sigma}}\tilde{\mathbf{V}}^{\top}
\]
with $\mathrm{rank}\left(\tilde{\mathbf{U}}\right)=\mathrm{rank}\left(\tilde{\mathbf{V}}\right)=\mathrm{rank}\left(\tilde{\boldsymbol{\Sigma}}\right)$,
we can construct an OOM $\mathcal{M}^{\prime}=(\boldsymbol{\omega}^{\prime},\{\boldsymbol{\Xi}^{\prime}(x)\}_{x\in\mathcal{O}},\boldsymbol{\sigma}^{\prime})$
with
\begin{eqnarray}
\boldsymbol{\omega}^{\prime} & = & \hat{\boldsymbol{\omega}}\left(\mathbf{G}_{\sigma}^{\top}\mathbf{V}\right)^{-1}\\
\boldsymbol{\Xi}^{\prime}(x) & = & \left(\mathbf{G}_{\sigma}^{\top}\mathbf{V}\right)\hat{\boldsymbol{\Xi}}\left(x\right)\left(\mathbf{G}_{\sigma}^{\top}\mathbf{V}\right)^{-1}\label{eq:Xi-x-prime}\\
\boldsymbol{\sigma}^{\prime} & = & \left(\mathbf{G}_{\sigma}^{\top}\mathbf{V}\right)\hat{\boldsymbol{\sigma}}\label{eq:sigma-prime}
\end{eqnarray}
which is obviously equivalent to $\hat{\mathcal{M}}$.

We can obtain from $\mathrm{rank}\left(\mathbf{U}\boldsymbol{\Sigma}\mathbf{V}^{\top}\right)=\mathrm{rank}\left(\mathbf{G}_{\omega}\mathbf{G}_{\sigma}^{\top}\right)=m$
that
\[
\left(\mathbf{U}\boldsymbol{\Sigma}\mathbf{V}^{\top}\right)^{+}=\mathbf{V}\boldsymbol{\Sigma}^{-1}\mathbf{U}^{\top}\stackrel{p}{\to}\left(\mathbf{G}_{\omega}\mathbf{G}_{\sigma}^{\top}\right)^{+}
\]
where $\mathbf{A}^{+}$ denotes the Moore-Penrose pseudoinverse of
$\mathbf{A}$, so
\begin{eqnarray*}
\boldsymbol{\omega}^{\prime} & = & \hat{\bar{\boldsymbol{\phi}}}_{2}^{\top}\mathbf{V}\left(\mathbf{G}_{\sigma}^{\top}\mathbf{V}\right)^{-1}\\
 & \stackrel{p}{\to} & \boldsymbol{\omega}^{*}(T_{0})\\
\boldsymbol{\Xi}^{\prime}(x) & = & \left(\mathbf{G}_{\sigma}^{\top}\mathbf{V}\right)\boldsymbol{\Sigma}^{-1}\mathbf{U}^{\top}\hat{\mathbf{C}}_{1,3}\left(x\right)\mathbf{V}\left(\mathbf{G}_{\sigma}^{\top}\mathbf{V}\right)^{-1}\\
 & \stackrel{p}{\to} & \mathbf{G}_{\sigma}^{\top}\mathbf{V}\boldsymbol{\Sigma}^{-1}\mathbf{U}^{\top}\mathbf{G}_{\omega}\boldsymbol{\Xi}(x)\\
 & \stackrel{p}{\to} & \mathbf{G}_{\sigma}^{\top}\left(\mathbf{G}_{\omega}\mathbf{G}_{\sigma}^{\top}\right)^{+}\mathbf{G}_{\omega}\boldsymbol{\Xi}(x)\\
 & = & \mathbf{G}_{\omega}^{+}\mathbf{G}_{\omega}\mathbf{G}_{\sigma}^{\top}\left(\mathbf{G}_{\omega}\mathbf{G}_{\sigma}^{\top}\right)^{+}\mathbf{G}_{\omega}\mathbf{G}_{\sigma}^{\top}\mathbf{G}_{\sigma}^{+\top}\boldsymbol{\Xi}(x)\\
 & = & \boldsymbol{\Xi}(x)\\
\boldsymbol{\sigma}^{\prime} & = & \mathbf{G}_{\sigma}^{\top}\mathbf{V}\boldsymbol{\Sigma}^{-1}\mathbf{U}^{\top}\hat{\bar{\boldsymbol{\phi}}}_{1}\\
 & \stackrel{p}{\to} & \boldsymbol{\sigma}
\end{eqnarray*}

Note $\boldsymbol{\omega}^{\prime}\stackrel{p}{\to}\boldsymbol{\omega}$
does not hold in general cases.

\paragraph{Part (2)}

We now consider the case of $I=I_{0}$ and $T\to\infty$.

According to Assumption \ref{assu:stationary}, the limit
\begin{eqnarray*}
\hat{\mathbf{C}}_{1,2} & \stackrel{p}{\to} & \mathbb{E}_{\infty}\left[\boldsymbol{\phi}_{1}(x_{t-L:t-1})\boldsymbol{\phi}_{2}(x_{t:t+L-1})^{\top}\right]\\
 & = & \lim_{k\to\infty}\sum_{z_{1:L}}\boldsymbol{\phi}_{1}(z_{1:L})\boldsymbol{\omega}\boldsymbol{\Xi}\left(\mathcal{O}\right)^{k}\boldsymbol{\Xi}(z_{1:L})\mathbf{G}_{\sigma}^{\top}
\end{eqnarray*}
exists. Then
\begin{eqnarray*}
\hat{\bar{\boldsymbol{\phi}}}_{1} & \stackrel{p}{\to} & \mathbb{E}_{\infty}\left[\boldsymbol{\phi}_{1}(x_{t-L:t-1})\right]=\mathbf{G}_{\omega}\boldsymbol{\sigma}\\
\hat{\bar{\boldsymbol{\phi}}}_{2}^{\top} & \stackrel{p}{\to} & \mathbb{E}_{\infty}\left[\boldsymbol{\phi}_{2}(x_{t:t+L-1})^{\top}\right]=\lim_{k\to\infty}\boldsymbol{\omega}\boldsymbol{\Xi}\left(\mathcal{O}\right)^{k}\mathbf{G}_{\sigma}^{\top}\\
\hat{\mathbf{C}}_{1,2} & \stackrel{p}{\to} & \mathbb{E}_{\infty}\left[\hat{\mathbf{C}}_{1,2}\right]=\mathbf{G}_{\omega}\mathbf{G}_{\sigma}^{\top}\\
\hat{\mathbf{C}}_{1,3}\left(x\right) & \stackrel{p}{\to} & \mathbb{E}_{\infty}\left[\hat{\mathbf{C}}_{1,2}\left(x\right)\right]=\mathbf{G}_{\omega}\boldsymbol{\Xi}(x)\mathbf{G}_{\sigma}^{\top}
\end{eqnarray*}
with
\begin{equation}
\mathbf{G}_{\omega}=\lim_{k\to\infty}\sum_{z_{1:L}}\boldsymbol{\phi}_{1}(z_{1:L})\boldsymbol{\omega}\boldsymbol{\Xi}\left(\mathcal{O}\right)^{k}\boldsymbol{\Xi}(z_{1:L})\label{eq:G-omega-infinite}
\end{equation}

The remaining part of the proof is omitted because it is the same
as in Part (1).

\subsection{Asymptotic correctness of nonequilibrium learning with different
initial states\label{subsec:Asymptotic-correctness-different-initial}}

If the $i$-th observation trajectories is generated by OOM $\mathcal{M}=(\boldsymbol{\omega}^{i},\{\boldsymbol{\Xi}(x)\}_{x\in\mathcal{O}},\boldsymbol{\sigma})$
for $i=1,\ldots,I$, and
\[
\boldsymbol{\omega}^{**}=\left\{ \begin{array}{ll}
\frac{1}{I}\sum_{i=1}^{I}\boldsymbol{\omega}^{i}, & \text{for }T\to\infty\\
\plim_{I\to\infty}\frac{1}{I}\sum_{i=1}^{I}\boldsymbol{\omega}^{i}, & \text{for }I\to\infty
\end{array}\right.
\]
the asymptotic correctness can also be shown as in Appendix \ref{subsec:Proof-of-Theorem-nonequilibrium}
by setting
\[
\mathbf{G}_{\omega}=\sum_{z_{1:L}}\boldsymbol{\phi}_{1}(z_{1:L})\boldsymbol{\omega}^{*}(T_{0})\boldsymbol{\Xi}(z_{1:L})
\]
with
\[
\boldsymbol{\omega}^{*}(T)=\frac{1}{T-2L}\boldsymbol{\omega}^{**}\sum_{t=1}^{T-2L}\boldsymbol{\Xi}(\mathcal{O})^{t-1}
\]
for $I\to\infty$, and
\[
\mathbf{G}_{\omega}=\lim_{k\to\infty}\sum_{z_{1:L}}\boldsymbol{\phi}_{1}(z_{1:L})\boldsymbol{\omega}^{**}\boldsymbol{\Xi}\left(\mathcal{O}\right)^{k}\boldsymbol{\Xi}(z_{1:L})
\]
for $T\to\infty$.

\subsection{Proof of Theorem \ref{thm:nonequilibrium-omega}\label{subsec:Proof-of-Corollary-nonequilibrium-omega}}

\paragraph{Part (1)}

We first show that there is an OOM $\mathcal{M}_{\mathrm{eq}}=(\boldsymbol{\omega}_{\mathrm{eq}},\{\boldsymbol{\Xi}(x)\}_{x\in\mathcal{O}},\boldsymbol{\sigma})$
which can describe the equilibrium dynamics of $\{x_{t}\}$.

In the case of $T=T_{0}$ and $I\to\infty$, we can obtain from Assumptions
\ref{assu:stationary} and \ref{assu:invertible} that
\begin{eqnarray}
\lim_{k\to\infty}\mathbf{G}_{\omega}\boldsymbol{\Xi}(\mathcal{O})^{k}\mathbf{G}_{\sigma}^{\top} & = & \lim_{k\to\infty}\frac{1}{T_{0}-2L}\sum_{t=0}^{T_{0}-2L-1}\mathbb{E}\left[\boldsymbol{\phi}_{1}\left(x_{t+1:t+L}\right)\boldsymbol{\phi}_{2}\left(x_{t+L+k+1:t+2L+k}\right)^{\top}\right]\nonumber \\
 & = & \left(\frac{1}{T_{0}-2L}\sum_{t=0}^{T_{0}-2L-1}\mathbb{E}\left[\boldsymbol{\phi}_{1}\left(x_{t+1:t+L}\right)\right]\right)\left(\mathbb{E}_{\infty}\left[\boldsymbol{\phi}_{2}\left(x_{t+1:t+L}\right)^{\top}\right]\right)\nonumber \\
 & = & \mathbf{G}_{\omega}\boldsymbol{\sigma}\left(\mathbb{E}_{\infty}\left[\boldsymbol{\phi}_{2}\left(x_{t+1:t+L}\right)^{\top}\right]\right)\nonumber \\
\Rightarrow\lim_{k\to\infty}\boldsymbol{\Xi}(\mathcal{O})^{k} & = & \boldsymbol{\sigma}\boldsymbol{\omega}_{\mathrm{eq}}\label{eq:XiO-k-limit-finite}
\end{eqnarray}
with
\begin{equation}
\boldsymbol{\omega}_{\mathrm{eq}}=\left(\mathbb{E}_{\infty}\left[\boldsymbol{\phi}_{2}\left(x_{t+1:t+L}\right)^{\top}\right]\right)\mathbf{G}_{\sigma}^{+\top}\label{eq:omega-bar}
\end{equation}
where $\mathbf{G}_{\omega}$ and $\mathbf{G}_{\sigma}$ are defined
by (\ref{eq:G-omega-finite}) and (\ref{eq:G-sigma}). Then 
\begin{eqnarray*}
\lim_{t\to\infty}\mathbb{P}\left(x_{t+1:t+l}=z_{1:l}\right) & = & \lim_{t\to\infty}\boldsymbol{\omega}\boldsymbol{\Xi}(\mathcal{O})^{t}\boldsymbol{\Xi}(z_{1:l})\boldsymbol{\sigma}\\
 & = & \boldsymbol{\omega}\boldsymbol{\Xi}(\mathcal{O})\boldsymbol{\sigma}\boldsymbol{\omega}_{\mathrm{eq}}\boldsymbol{\Xi}(z_{1:l})\boldsymbol{\sigma}\\
 & = & \boldsymbol{\omega}_{\mathrm{eq}}\boldsymbol{\Xi}(z_{1:l})\boldsymbol{\sigma}
\end{eqnarray*}

In the case of $I=I_{0}$ and $T\to\infty$, because $\mathrm{rank}\left(\mathbf{G}_{\omega}\right)=m$
for $\mathbf{G}_{\omega}$ defined by (\ref{eq:G-omega-infinite}),
there is a sufficiently large but finite $T^{\prime}$ so that $\mathrm{rank}\left(\mathbf{G}_{\omega}^{\prime}\right)=m$
with
\[
\mathbf{G}_{\omega}^{\prime}=\sum_{z_{1:L}}\boldsymbol{\phi}_{1}(z_{1:L})\boldsymbol{\omega}\boldsymbol{\Xi}\left(\mathcal{O}\right)^{T^{\prime}}\boldsymbol{\Xi}(z_{1:L})
\]
Considering
\begin{eqnarray}
\lim_{k\to\infty}\mathbf{G}_{\omega}^{\prime}\boldsymbol{\Xi}(\mathcal{O})^{k}\mathbf{G}_{\sigma}^{\top} & = & \lim_{k\to\infty}\mathbb{E}\left[\boldsymbol{\phi}_{1}\left(x_{T^{\prime}+1:T^{\prime}+L}\right)\boldsymbol{\phi}_{2}\left(x_{T^{\prime}+L+k+1:T^{\prime}+2L+k}\right)^{\top}\right]\nonumber \\
 & = & \mathbf{G}_{\omega}^{\prime}\boldsymbol{\sigma}\left(\mathbb{E}_{\infty}\left[\boldsymbol{\phi}_{2}\left(x_{t+1:t+L}\right)^{\top}\right]\right)\nonumber \\
\Rightarrow\lim_{k\to\infty}\boldsymbol{\Xi}(\mathcal{O})^{k} & = & \boldsymbol{\sigma}\boldsymbol{\omega}_{\mathrm{eq}}\label{eq:XiO-k-limit-infinite}
\end{eqnarray}
with $\boldsymbol{\omega}_{\mathrm{eq}}$ defined by (\ref{eq:omega-bar}),
we can also conclude that
\[
\lim_{t\to\infty}\mathbb{P}\left(x_{t+1:t+l}=z_{1:l}\right)=\boldsymbol{\omega}_{\mathrm{eq}}\boldsymbol{\Xi}(z_{1:l})\boldsymbol{\sigma}
\]

Note in both cases, $\boldsymbol{\omega}_{\mathrm{eq}}$ satisfies
$\boldsymbol{\omega}_{\mathrm{eq}}\lim_{k\to\infty}\boldsymbol{\Xi}(\mathcal{O})^{k}=\boldsymbol{\omega}_{\mathrm{eq}}$
and 
\begin{eqnarray*}
\boldsymbol{\omega}_{\mathrm{eq}}\boldsymbol{\Xi}(\mathcal{O}) & = & \lim_{t\to\infty}\boldsymbol{\omega}_{\mathrm{eq}}\boldsymbol{\Xi}(\mathcal{O})^{t+1}\\
 & = & \boldsymbol{\omega}_{\mathrm{eq}}\\
\boldsymbol{\omega}_{\mathrm{eq}}\boldsymbol{\sigma} & = & \boldsymbol{\omega}_{\mathrm{eq}}\boldsymbol{\Xi}(\mathcal{O})\boldsymbol{\sigma}\\
 & = & \lim_{t\to\infty}\sum_{x\in\mathcal{O}}\mathbb{P}\left(x_{t}=x\right)=1
\end{eqnarray*}

\paragraph{Part (2)}

In this part, we show that
\[
\mathbf{w}\boldsymbol{\Xi}(\mathcal{O})=\mathbf{w},\quad\mathbf{w}\boldsymbol{\sigma}=1
\]
has a unique solution $\mathbf{w}=\boldsymbol{\omega}_{\mathrm{eq}}$.

According to Appendix \ref{subsec:Proof-of-Theorem-nonequilibrium}
and (\ref{eq:XiO-k-limit-finite}), (\ref{eq:XiO-k-limit-infinite}),
if $\mathbf{w}\boldsymbol{\Xi}(\mathcal{O})=\mathbf{w}$ and $\mathbf{w}\boldsymbol{\sigma}=1$,
we have
\begin{eqnarray*}
\mathbf{w} & = & \lim_{k\to\infty}\mathbf{w}\boldsymbol{\Xi}(\mathcal{O})^{k}\\
 & = & \mathbf{w}\boldsymbol{\sigma}\boldsymbol{\omega}_{\mathrm{eq}}\\
 & = & \boldsymbol{\omega}_{\mathrm{eq}}
\end{eqnarray*}

\paragraph{Part (3)}

We now show Theorem \ref{thm:nonequilibrium-omega}.

The problem (\ref{eq:nonequilibrium-omega-opt}) is equivalent to
\begin{eqnarray*}
\min_{\mathbf{w}^{\prime}}E\left(\mathbf{w}^{\prime}\right) & = & \left(\mathbf{w}^{\prime}\boldsymbol{\Xi}^{\prime}\left(\mathcal{O}\right)-\mathbf{w}^{\prime}\right)\left(\mathbf{G}_{\sigma}^{\top}\mathbf{V}\right)\\
 &  & \cdot\left(\mathbf{G}_{\sigma}^{\top}\mathbf{V}\right)^{\top}\left(\mathbf{w}^{\prime}\boldsymbol{\Xi}^{\prime}\left(\mathcal{O}\right)-\mathbf{w}^{\prime}\right)^{\top}\\
\mathrm{s.t.} &  & \mathbf{w}^{\prime}\boldsymbol{\sigma}^{\prime}=1
\end{eqnarray*}
where $\boldsymbol{\Xi}^{\prime}\left(\mathcal{O}\right)=\sum_{x\in\mathcal{O}}\boldsymbol{\Xi}^{\prime}\left(x\right)$,
$\boldsymbol{\Xi}^{\prime}\left(x\right)$ and $\boldsymbol{\sigma}^{\prime}$
are given by (\ref{eq:Xi-x-prime}) and (\ref{eq:sigma-prime}), and
$\mathbf{w}^{\prime}$ is related to $\mathbf{w}$ with $\mathbf{w}^{\prime}=\mathbf{w}\left(\mathbf{G}_{\sigma}^{\top}\mathbf{V}\right)^{-1}$.
This problem can be further transformed into an unconstrained one
\begin{equation}
\min_{\mathbf{w}^{\prime}}E\left(\mathbf{w}^{\prime}\left(\mathbf{I}-\boldsymbol{\sigma}^{\prime}\boldsymbol{\sigma}^{\prime+}\right)+\boldsymbol{\sigma}^{\prime+}\right)+\left\Vert \mathbf{w}^{\prime}\left(\mathbf{I}-\boldsymbol{\sigma}^{\prime}\boldsymbol{\sigma}^{\prime+}\right)+\boldsymbol{\sigma}^{\prime+}-\mathbf{w}^{\prime}\right\Vert ^{2}\label{eq:omega-eq-unconstraint}
\end{equation}
where $\mathbf{w}^{\prime}\left(\mathbf{I}-\boldsymbol{\sigma}^{\prime}\boldsymbol{\sigma}^{\prime+}\right)+\boldsymbol{\sigma}^{\prime+}$
is the projection of $\mathbf{w}^{\prime}$ on the space $\{\mathbf{w}^{\prime}|\mathbf{w}^{\prime}\boldsymbol{\sigma}^{\prime}=1\}$
and $\mathbf{I}$ denotes the identity matrix of appropriate dimension.
Considering that $\boldsymbol{\Xi}^{\prime}\left(x\right)\stackrel{p}{\to}\boldsymbol{\Xi}\left(x\right)$,
$\boldsymbol{\sigma}^{\prime}\stackrel{p}{\to}\boldsymbol{\sigma}$,
\begin{eqnarray*}
\left(\mathbf{G}_{\sigma}^{\top}\mathbf{V}\right)\left(\mathbf{G}_{\sigma}^{\top}\mathbf{V}\right)^{\top} & = & \mathbf{G}_{\sigma}^{\top}\mathbf{V}\boldsymbol{\Sigma}^{-1}\mathbf{U}^{\top}\mathbf{U}\boldsymbol{\Sigma}\mathbf{V}^{\top}\mathbf{G}_{\sigma}\\
 & \stackrel{p}{\to} & \mathbf{G}_{\sigma}^{\top}\left(\mathbf{G}_{\omega}^{\top}\mathbf{G}_{\sigma}\right)^{+}\mathbf{G}_{\omega}\mathbf{G}_{\sigma}^{\top}\mathbf{G}_{\sigma}\\
 & = & \mathbf{G}_{\sigma}^{\top}\mathbf{G}_{\sigma}
\end{eqnarray*}
and the conclusion in Part (2), we can obtain that the optimal solution
of (\ref{eq:omega-eq-unconstraint}) converges to $\boldsymbol{\omega}_{\mathrm{eq}}$
in probability and $\hat{\boldsymbol{\omega}}_{\mathrm{eq}}\stackrel{p}{\to}\boldsymbol{\omega}_{\mathrm{eq}}\left(\mathbf{G}_{\sigma}^{\top}\mathbf{V}\right)^{-1}$
according to Theorem 2.7 in {[}1{]}, which yields the conclusion
of Theorem \ref{thm:nonequilibrium-omega}.

\paragraph{Part (4)}

We derive in this part the closed-form solution to (\ref{eq:nonequilibrium-omega-opt}).

Since the projection of $\mathbf{w}^{\prime\prime}$ on the space
$\{\mathbf{w}^{\prime\prime}|\mathbf{w}^{\prime\prime}\hat{\boldsymbol{\sigma}}=1\}$
is $\mathbf{w}^{\prime\prime}\left(\mathbf{I}-\hat{\boldsymbol{\sigma}}\hat{\boldsymbol{\sigma}}^{+}\right)+\hat{\boldsymbol{\sigma}}^{+}$,
(\ref{eq:nonequilibrium-omega-opt}) can be equivalent transformed
into
\[
\min_{\mathbf{w}^{\prime\prime}}\left\Vert \mathbf{w}^{\prime\prime}\left(\mathbf{I}-\hat{\boldsymbol{\sigma}}\hat{\boldsymbol{\sigma}}^{+}\right)\left(\hat{\boldsymbol{\Xi}}(\mathcal{O})-\mathbf{I}\right)+\hat{\boldsymbol{\sigma}}^{+}\left(\hat{\boldsymbol{\Xi}}(\mathcal{O})-\mathbf{I}\right)\right\Vert ^{2}
\]
The solution to this problem is
\[
\mathbf{w}^{*}=-\hat{\boldsymbol{\sigma}}^{+}\left(\hat{\boldsymbol{\Xi}}(\mathcal{O})-\mathbf{I}\right)\left(\left(\mathbf{I}-\hat{\boldsymbol{\sigma}}\hat{\boldsymbol{\sigma}}^{+}\right)\left(\hat{\boldsymbol{\Xi}}(\mathcal{O})-\mathbf{I}\right)\right)^{+}
\]
which provides the optimal value of $\hat{\boldsymbol{\omega}}_{\mathrm{eq}}$
as
\begin{eqnarray}
\hat{\boldsymbol{\omega}}_{\mathrm{eq}} & = & \mathbf{w}^{*}\left(\mathbf{I}-\hat{\boldsymbol{\sigma}}\hat{\boldsymbol{\sigma}}^{+}\right)+\hat{\boldsymbol{\sigma}}^{+}\nonumber \\
 & = & \hat{\boldsymbol{\sigma}}^{+}-\hat{\boldsymbol{\sigma}}^{+}\left(\hat{\boldsymbol{\Xi}}(\mathcal{O})-\mathbf{I}\right)\left(\left(\mathbf{I}-\hat{\boldsymbol{\sigma}}\hat{\boldsymbol{\sigma}}^{+}\right)\left(\hat{\boldsymbol{\Xi}}(\mathcal{O})-\mathbf{I}\right)\right)^{+}\left(\mathbf{I}-\hat{\boldsymbol{\sigma}}\hat{\boldsymbol{\sigma}}^{+}\right)\label{eq:closed-form-solution}
\end{eqnarray}

\subsection{Proof of Theorem \ref{thm:bloom}}

Here we only consider the consistency of the binless OOM as $I\to\infty$.
The proof can be easily to extended to the case of $T\to\infty$.
In addition, we denote $\mathbb{E}_{\infty}[g(x_{t+1:t+r})]$ and
$\mathbb{E}[g(x_{1:r})|\hat{\mathcal{M}}_{\mathrm{eq}}]$ by $\mathbb{E}_{\infty}[g]$
and $\mathbb{E}_{\hat{\mathcal{M}}}[g]$ for convenience of notation.

\paragraph{Part (1)}

We first show that Theorem \ref{thm:bloom} holds for $g\left(x_{t+1:t+r}\right)=1_{x_{t+1:t+r}\in\mathcal{B}_{i_{1}}\times\mathcal{B}_{i_{2}}\times\ldots\times\mathcal{B}_{i_{r}}}$,
where $\mathcal{B}_{1},\ldots,\mathcal{B}_{K}$ is a partition of
$\mathcal{O}$ and $i_{1:r}\in\{1,\ldots,K\}^{r}$. In this case,
we can construct a discrete OOM with observation space $\{\mathcal{B}_{1},\ldots,\mathcal{B}_{K}\}$
by the nonequilibrium learning algorithm, which can provide the same
estimate of $\mathbb{E}_{\infty}\left[g\left(x_{t+1:t+r}\right)\right]$
as $\hat{\mathcal{M}}_{\mathrm{eq}}$. Therefore, we can show $\mathbb{E}_{\hat{\mathcal{M}}}[g]\stackrel{p}{\to}\mathbb{E}_{\infty}[g]$
by using the similar proof of Theorem \ref{thm:nonequilibrium-omega}.

\paragraph{Part (2)}

We now consider the case that $g$ is a continuous function. According
to the Heine-Cantor theorem, $g$ is also uniformly continuous. Then,
for an arbitrary $\epsilon>0$, we can construct a simple function
\[
\hat{g}(x_{t+1:t+r})=\sum_{i_{1},\ldots,i_{r}}c_{i_{1}i_{2}\ldots i_{r}}1_{x_{t+1:t+r}\in\mathcal{B}_{i_{1}}\times\ldots\times\mathcal{B}_{i_{r}}}
\]
so that
\[
\left|g(z_{1:r})-\hat{g}(z_{1:r})\right|\le\epsilon,\quad\forall z_{1:r}\in\mathcal{O}^{r}
\]
where $\{\mathcal{B}_{1},\ldots,\mathcal{B}_{K}\}$ is a partition
of $\mathcal{O}$. Then, we have
\[
\left|\mathbb{E}_{\infty}[g]-\mathbb{E}_{\infty}[\hat{g}]\right|\le\mathbb{E}_{\infty}[\left|g-\hat{g}\right|]\le\epsilon
\]
and
\[
\left|\mathbb{E}_{\infty}[\hat{g}]-\mathbb{E}_{\hat{\mathcal{M}}}[\hat{g}]\right|\stackrel{p}{\to}0
\]
as $I\to\infty$ according to the conclusion of Part (1), where $\mathbb{E}_{\infty}[g]=\mathbb{E}_{\infty}[g(x_{t+1:t+r})]$
and $\mathbb{E}_{\hat{\mathcal{M}}}[g]=\mathbb{E}[g(x_{1:r})|\hat{\mathcal{M}}_{\mathrm{eq}}]$.

It can be known from the boundness of feature functions, there exists
a constant $\xi$ such that
\begin{equation}
1_{\max_{x\in\mathcal{X}}\left\Vert \hat{\mathbf{W}}_{x}\right\Vert <\xi/\left|\mathcal{X}\right|}\stackrel{p}{\to}1\label{eq:Xi-bound}
\end{equation}
Under the condition that $\max_{x\in\mathcal{X}}\left\Vert \hat{\mathbf{W}}_{x}\right\Vert <\xi/\left|\mathcal{X}\right|$,
we have

\begin{eqnarray*}
\left|\mathbb{E}_{\hat{\mathcal{M}}_{\mathrm{eq}}}[\hat{g}]-\mathbb{E}_{\hat{\mathcal{M}}_{\mathrm{eq}}}[g]\right| & = & \hat{\boldsymbol{\omega}}_{\mathrm{eq}}\left(\sum_{z_{1:r}\in\mathcal{X}^{r}}\left(\hat{g}(z_{1:r})-g(z_{1:r})\right)\mathbf{W}_{z_{1}}\ldots\mathbf{W}_{z_{r}}\right)\hat{\boldsymbol{\sigma}}\\
 & \le & \left\Vert \hat{\boldsymbol{\omega}}_{\mathrm{eq}}\right\Vert \left\Vert \hat{\boldsymbol{\sigma}}\right\Vert \left(\sum_{z_{1:r}\in\mathcal{X}^{r}}\frac{\xi^{r}\epsilon}{\left|\mathcal{X}\right|^{r}}\right)\\
 & = & \left\Vert \hat{\boldsymbol{\omega}}_{\mathrm{eq}}\right\Vert \left\Vert \hat{\boldsymbol{\sigma}}\right\Vert \xi^{r}\epsilon
\end{eqnarray*}
In addition, considering that we can show as in Appendix \ref{subsec:Proof-of-Theorem-nonequilibrium}
that
\begin{eqnarray*}
\hat{\boldsymbol{\omega}}_{\mathrm{eq}} & \stackrel{p}{\to} & \boldsymbol{\omega}_{\mathrm{eq}}\mathbf{G}_{\sigma}^{\top}\mathbf{V}\\
\hat{\boldsymbol{\sigma}} & \stackrel{p}{\to} & \left(\mathbf{G}_{\sigma}^{\top}\mathbf{V}\right)^{-1}\boldsymbol{\sigma}
\end{eqnarray*}
we can obtain
\begin{equation}
1_{\left\Vert \hat{\boldsymbol{\omega}}_{\mathrm{eq}}\right\Vert \left\Vert \hat{\boldsymbol{\sigma}}\right\Vert \le\xi_{0}}\stackrel{p}{\to}1\label{eq:omega-sigma-bound}
\end{equation}
and
\[
1_{\left|\mathbb{E}_{\hat{\mathcal{M}}_{\mathrm{eq}}}[\hat{g}]-\mathbb{E}_{\hat{\mathcal{M}}_{\mathrm{eq}}}[g]\right|\le\xi_{0}\xi^{r}\epsilon}\stackrel{p}{\to}1
\]
where $\xi_{0}$ is a constant larger than $\left\Vert \hat{\boldsymbol{\omega}}_{\mathrm{eq}}\right\Vert \cdot\left\Vert \hat{\boldsymbol{\sigma}}\right\Vert $.

Based on the above analysis and the fact that
\begin{eqnarray*}
\left|\mathbb{E}_{\infty}[g]-\mathbb{E}_{\hat{\mathcal{M}}_{\mathrm{eq}}}[g]\right| & = & \left|\mathbb{E}_{\infty}[g]-\mathbb{E}_{\infty}[\hat{g}]+\mathbb{E}_{\infty}[\hat{g}]-\mathbb{E}_{\hat{\mathcal{M}}_{\mathrm{eq}}}[\hat{g}]+\mathbb{E}_{\hat{\mathcal{M}}_{\mathrm{eq}}}[\hat{g}]-\mathbb{E}_{\hat{\mathcal{M}}_{\mathrm{eq}}}[g]\right|\\
 & \le & \left|\mathbb{E}_{\infty}[g]-\mathbb{E}_{\infty}[\hat{g}]\right|+\left|\mathbb{E}_{\infty}[\hat{g}]-\mathbb{E}_{\hat{\mathcal{M}}_{\mathrm{eq}}}[\hat{g}]\right|+\left|\mathbb{E}_{\hat{\mathcal{M}}_{\mathrm{eq}}}[\hat{g}]-\mathbb{E}_{\hat{\mathcal{M}}_{\mathrm{eq}}}[g]\right|
\end{eqnarray*}
we can get
\begin{eqnarray*}
\Pr\left(\left|\mathbb{E}_{\infty}[g]-\mathbb{E}_{\hat{\mathcal{M}}_{\mathrm{eq}}}[g]\right|\le\left(\xi_{0}\xi^{r}+2\right)\epsilon\right) & \ge & \Pr\left(\left|\mathbb{E}_{\infty}[g]-\mathbb{E}_{\infty}[\hat{g}]\right|\le\epsilon,\left|\mathbb{E}_{\infty}[\hat{g}]-\mathbb{E}_{\hat{\mathcal{M}}_{\mathrm{eq}}}[\hat{g}]\right|\le\epsilon,\right.\\
 &  & \left.\left|\mathbb{E}_{\hat{\mathcal{M}}_{\mathrm{eq}}}[\hat{g}]-\mathbb{E}_{\hat{\mathcal{M}}_{\mathrm{eq}}}[g]\right|\le\xi_{0}\xi^{r}\epsilon\right)\\
 & \to & 1
\end{eqnarray*}
Because this equation holds for all $\epsilon>0$, we can conclude
that $\mathbb{E}_{\hat{\mathcal{M}}_{\mathrm{eq}}}[g]\stackrel{p}{\to}\mathbb{E}_{\infty}[g]$.

\paragraph{Part (3)}

In this part, we prove the conclusion of the theorem in the case where
$g$ is a Borel measurable function and bounded with $|g(z_{1:r})|<\xi_{g}$
for all $z_{1:r}\in\mathcal{O}^{r}$, and there exist constants $\bar{\xi}$
and $\underline{\xi}$ so that $\left\Vert \boldsymbol{\Xi}\left(x\right)\right\Vert \le\bar{\xi}$
and $\lim_{t\to\infty}\mathbb{P}\left(x_{t+1:t+r}=z_{1:r}\right)\ge\underline{\xi}$
for all $x\in\mathcal{O}$ and $z_{1:r}\in\mathcal{O}^{r}$.

According to Theorem 2.2 in {[}2{]}, for an arbitrary $\epsilon>0$,
there is a continuous function $\hat{g}^{\prime}$ satisfies $\mathbb{E}_{\infty}[1_{x_{t+1:t+r}\in\mathcal{K}_{\epsilon}(\hat{g}^{\prime})}]<\epsilon$,
where $\mathcal{K}_{\epsilon}(\hat{g}^{\prime})=\{z_{1:r}|z_{1:r}\in\mathcal{O}^{r},\left|\hat{g}^{\prime}(z_{1:r})-g(z_{1:r})\right|>\epsilon\}$.
Define
\[
\hat{g}(z_{1:r})=\left\{ \begin{array}{ll}
\hat{g}^{\prime}(z_{1:r}), & \left|\hat{g}^{\prime}(z_{1:r})\right|\le\xi_{g}\\
-\xi_{g}, & \hat{g}^{\prime}(z_{1:r})<-\xi_{g}\\
\xi_{g}, & \hat{g}^{\prime}(z_{1:r})>\xi_{g}
\end{array}\right.
\]

It can be seen that $\hat{g}$ is a continuous function which is also
satisfies $\mathbb{E}_{\infty}[1_{x_{t+1:t+r}\in\mathcal{K}_{\epsilon}(\hat{g})}]<\epsilon$
and bounded with $|\hat{g}(z_{1:r})|<\xi_{g}$. So the difference
between $\mathbb{E}_{\infty}[g]$ and $\mathbb{E}_{\infty}[\hat{g}]$
satisfies
\begin{eqnarray*}
\left|\mathbb{E}_{\infty}[g]-\mathbb{E}_{\infty}[\hat{g}]\right| & \le & \mathbb{E}_{\infty}\left[\left|g(x_{t+1:t+r})-\hat{g}(x_{t+1:t+r})\right|\right]\\
 & = & \mathbb{E}_{\infty}[1_{x_{t+1:t+r}\in\mathcal{K}_{\epsilon}(\hat{g})}]\mathbb{E}_{\infty}\left[\left|g(x_{t+1:t+r})-\hat{g}(x_{t+1:t+r})\right||x_{t+1:t+r}\in\mathcal{K}_{\epsilon}(\hat{g})\right]\\
 &  & +\mathbb{E}_{\infty}[1_{x_{t+1:t+r}\notin\mathcal{K}_{\epsilon}(\hat{g})}]\mathbb{E}_{\infty}\left[\left|g(x_{t+1:t+r})-\hat{g}(x_{t+1:t+r})\right||x_{t+1:t+r}\notin\mathcal{K}_{\epsilon}(\hat{g})\right]\\
 & \le & \epsilon\cdot2\xi_{g}+\epsilon=\left(2\xi_{g}+1\right)\epsilon
\end{eqnarray*}
For the difference between $\mathbb{E}_{\infty}[\hat{g}]$ and $\mathbb{E}_{\hat{\mathcal{M}}_{\mathrm{eq}}}[\hat{g}]$,
we can obtain from the above that $\left|\mathbb{E}_{\infty}[\hat{g}]-\mathbb{E}_{\hat{\mathcal{M}}_{\mathrm{eq}}}[\hat{g}]\right|\stackrel{p}{\to}0$
as $I\to\infty$ by considering that $\hat{g}$ is continuous, which
implies that there is an $I_{0}$ such that
\[
\Pr\left(\left|\mathbb{E}_{\infty}[\hat{g}]-\mathbb{E}_{\hat{\mathcal{M}}_{\mathrm{eq}}}[\hat{g}]\right|>\epsilon\right)<\epsilon,\quad\forall I>I_{0}
\]

Next, let us consider the value of $\left|\mathbb{E}_{\hat{\mathcal{M}}_{\mathrm{eq}}}[\hat{g}]-\mathbb{E}_{\hat{\mathcal{M}}_{\mathrm{eq}}}[g]\right|$.
Note that
\begin{eqnarray*}
\left|\mathbb{E}_{\hat{\mathcal{M}}}[\hat{g}]-\mathbb{E}_{\hat{\mathcal{M}}}[g]\right| & \le & \left\Vert \hat{\boldsymbol{\omega}}_{0}\right\Vert \left\Vert \hat{\boldsymbol{\sigma}}\right\Vert \left\Vert \sum_{z_{1:n}\in\mathcal{X}^{r}}\left(\hat{g}(z_{1:r})-g(z_{1:r})\right)\hat{\mathbf{W}}_{z_{1}}\ldots\hat{\mathbf{W}}_{z_{r}}\right\Vert \\
 & < & \frac{\xi_{0}\xi^{r}}{\left|\mathcal{X}\right|^{r}}\left|\sum_{z_{1:r}\in\mathcal{X}^{r}}\left(\hat{g}(z_{1:r})-g(z_{1:r})\right)\right|
\end{eqnarray*}
under the condition that $\left\Vert \hat{\mathbf{W}}_{x}\right\Vert <\xi/\left|\mathcal{X}\right|$
and $\left\Vert \hat{\boldsymbol{\omega}}_{\mathrm{eq}}\right\Vert \left\Vert \hat{\boldsymbol{\sigma}}\right\Vert \le\xi_{0}$.
Therefore, there exists an $I_{1}$ such that
\begin{equation}
\Pr\left(\left|\mathbb{E}_{\hat{\mathcal{M}}_{\mathrm{eq}}}[\hat{g}]-\mathbb{E}_{\hat{\mathcal{M}}_{\mathrm{eq}}}[g]\right|\ge\frac{\xi_{0}\xi^{r}}{\left|\mathcal{X}\right|^{r}}\left|\sum_{z_{1:r}\in\mathcal{X}^{r}}\left(\hat{g}(z_{1:r})-g(z_{1:r})\right)\right|\right)<\epsilon,\quad\forall I>I_{1}\label{eq:tmp-bound-1}
\end{equation}
due to (\ref{eq:Xi-bound}) and (\ref{eq:omega-sigma-bound}). Let
$x_{1:r}^{\prime}$ denotes a random sample taken uniformly from $\mathcal{X}^{r}$.
We can obtain that
\begin{eqnarray*}
\mathbb{P}\left(x_{1:r}^{\prime}\right) & = & \mathbb{P}\left(x_{1}^{\prime}\right)\ldots\mathbb{P}\left(x_{r}^{\prime}\right)\\
 & \le & \left(\left\Vert \boldsymbol{\omega}\right\Vert \left\Vert \boldsymbol{\sigma}\right\Vert \xi_{O}\bar{\xi}\right)^{r}
\end{eqnarray*}
where $\xi_{O}\ge\left\Vert \boldsymbol{\Xi}\left(\mathcal{O}\right)^{k}\right\Vert $
for any $k\ge0$. Note $\xi_{O}<\infty$ because we can show the existing
of the limit of $\{\left\Vert \boldsymbol{\Xi}\left(\mathcal{O}\right)^{0}\right\Vert ,\left\Vert \boldsymbol{\Xi}\left(\mathcal{O}\right)^{1}\right\Vert ,\ldots\}$
by similar steps in Appendix \ref{subsec:Proof-of-Corollary-nonequilibrium-omega}.
Thus 
\begin{eqnarray*}
\mathbb{E}\left[\frac{1}{\left|\mathcal{X}\right|^{r}}\left|\sum_{z_{1:r}\in\mathcal{X}^{r}}\left(\hat{g}(z_{1:r})-g(z_{1:r})\right)\right|\right] & \le & \mathbb{E}\left[\mathbb{E}\left[\left|\hat{g}(x_{1:r}^{\prime})-g(x_{1:r}^{\prime})\right||\mathcal{X}\right]\right]\\
 & = & \mathbb{E}\left[\left|\hat{g}(x_{1:r}^{\prime})-g(x_{1:r}^{\prime})\right|\right]\\
 & = & \mathbb{E}\left[1_{x_{1:r}^{\prime}\in\mathcal{K}_{\epsilon}(\hat{g})}\right]\mathbb{E}\left[\left|\hat{g}(x_{1:r}^{\prime})-g(x_{1:r}^{\prime})\right||x_{1:r}^{\prime}\in\mathcal{K}_{\epsilon}(\hat{g})\right]\\
 &  & +\mathbb{E}\left[1_{x_{1:r}^{\prime}\notin\mathcal{K}_{\epsilon}(\hat{g})}\right]\mathbb{E}\left[\left|\hat{g}(x_{1:r}^{\prime})-g(x_{1:r}^{\prime})\right||x_{1:r}^{\prime}\notin\mathcal{K}_{\epsilon}(\hat{g})\right]\\
 & \le & \xi_{\mu}\epsilon\cdot2\xi_{g}+\epsilon=\left(2\xi_{g}\xi_{\mu}+1\right)\epsilon
\end{eqnarray*}
where $\xi_{\mu}=\left(\left\Vert \boldsymbol{\omega}\right\Vert \left\Vert \boldsymbol{\sigma}\right\Vert \xi_{O}\bar{\xi}\right)^{r}/\underline{\xi}$.
By the Markov's inequality, we have
\begin{equation}
\Pr\left[\frac{1}{\left|\mathcal{X}\right|^{r}}\left|\sum_{z_{1:r}\in\mathcal{X}^{r}}\left(\hat{g}(z_{1:r})-g(z_{1:r})\right)\right|\ge\sqrt{\epsilon}\right]\le\left(2\xi_{g}\xi_{\mu}+1\right)\sqrt{\epsilon}\label{eq:tmp-bound-2}
\end{equation}
Combining (\ref{eq:tmp-bound-1}) and (\ref{eq:tmp-bound-2}) leads
to
\begin{eqnarray*}
\Pr\left(\left|\mathbb{E}_{\hat{\mathcal{M}}_{\mathrm{eq}}}[\hat{g}]-\mathbb{E}_{\hat{\mathcal{M}}_{\mathrm{eq}}}[g]\right|\ge\xi_{0}\xi^{r}\sqrt{\epsilon}\right) & \le & \Pr\left(\left|\mathbb{E}_{\hat{\mathcal{M}}_{\mathrm{eq}}}[\hat{g}]-\mathbb{E}_{\hat{\mathcal{M}}_{\mathrm{eq}}}[g]\right|\ge\frac{\xi_{0}\xi^{r}}{\left|\mathcal{X}\right|^{r}}\left|\sum_{z_{1:r}\in X^{r}}\left(\hat{g}(z_{1:r})-g(z_{1:r})\right)\right|\right)\\
 &  & +\Pr\left(\frac{1}{\left|\mathcal{X}\right|^{r}}\left|\sum_{z_{1:r}\in\mathcal{X}^{r}}\left(\hat{g}(z_{1:r})-g(z_{1:r})\right)\right|\ge\sqrt{\epsilon}\right)\\
 & \le & \epsilon+\left(2\xi_{g}\xi_{\mu}+1\right)\sqrt{\epsilon}
\end{eqnarray*}
for all $I>I_{1}$.

From all the above, we have
\begin{align*}
\hspace{2em} & \hspace{-2em}\Pr\left(\left|\mathbb{E}_{\infty}[g]-\mathbb{E}_{\hat{\mathcal{M}}_{\mathrm{eq}}}[g]\right|\le2(\xi_{g}+1)\epsilon+\xi_{0}\xi^{r}\sqrt{\epsilon}\right)\\
 & \ge\Pr\left(\left|\mathbb{E}_{\infty}[\hat{g}]-\mathbb{E}_{\hat{\mathcal{M}}_{\mathrm{eq}}}[\hat{g}]\right|\le\epsilon,\left|\mathbb{E}_{\hat{\mathcal{M}}_{\mathrm{eq}}}[\hat{g}]-\mathbb{E}_{\hat{\mathcal{M}}_{\mathrm{eq}}}[g]\right|\le\xi_{0}\xi^{r}\sqrt{\epsilon}\right)\\
 & \ge1-\Pr\left(\left|\mathbb{E}_{\infty}[\hat{g}]-\mathbb{E}_{\hat{\mathcal{M}}_{\mathrm{eq}}}[\hat{g}]\right|>\epsilon\right)-\Pr\left(\left|\mathbb{E}_{\hat{\mathcal{M}}_{\mathrm{eq}}}[\hat{g}]-\mathbb{E}_{\hat{\mathcal{M}}_{\mathrm{eq}}}[g]\right|>\xi_{0}\xi^{r}\sqrt{\epsilon}\right)\\
 & \ge1-2\epsilon-\left(2\xi_{g}\xi_{\mu}+1\right)\sqrt{\epsilon}
\end{align*}
for all $I>\max\{I_{0},I_{1}\}$, which yields $\mathbb{E}_{\hat{\mathcal{M}}_{\mathrm{eq}}}[g]\stackrel{p}{\to}\mathbb{E}_{\infty}[g]$
due to the arbitrariness of $\epsilon$.

\section{Settings in applications\label{sec:Settings-in-applications}}

\subsection{Models}

The one-dimensional diffusion processes in Section \ref{sec:Applications}
are driven by the Brownian dynamics with $\beta=0.3$, 
\[
V\left(x\right)=\frac{\sum_{i=1}^{5}\left(\left|x-c_{i}\right|+0.001\right)^{-2}u_{i}}{\sum_{i=1}^{5}\left(\left|x-c_{i}\right|+0.001\right)^{-2}}
\]
and the sample interval is $0.002$. For the two-dimensional process,
$\beta=2$, 
\[
V\left(x\right)=-\log\left(\sum_{i=1}^{3}p_{i}\mathcal{N}\left(x|\mu_{i},\Sigma_{i}\right)\right)
\]
and the sample interval is $0.01$, where $c_{1:5}=(-0.3,0.5,1,1.5,2.3)$,
$u_{1:5}=(21,4,8,-1,20)$, $p_{1:3}=(0.25,0.25,0.5)$, $\mu_{1}=(0,-0.5)$,
$\mu_{2}=(-1,0.5)$, $\mu_{3}=(1,-0.5)$. The simulation details of
alanine dipeptide is given in {[}3{]}.

\subsection{Algorithms}

The parameters of discrete spectral learning are chosen as: $L=3$,
$m=10$, and $\boldsymbol{\phi}_{1}=\boldsymbol{\phi}_{2}$ are indicator
functions of all $\mathcal{O}^{L}$ observation subsequences with
length $L$.

The parameters of binless spectral learning are almost the same as
discrete ones, except $\boldsymbol{\phi}_{1}=\boldsymbol{\phi}_{2}$
are Gaussian activation functions with random weights of functional
link neural networks with $D_{1}=D_{2}=100$.

The number of hidden states of HMMs is $10$. For continuous data,
we partition the state space into $100$ discrete bins $k$-mean clustering,
and then learn HMMs by the EM algorithm, where the HMM package in
PyEMMA {[}4{]} is used. All observation samples within the same bin
are assumed to be independent for quantitative analysis. 

\section*{References}
\begin{enumerate}
\item[{{[}1{]}}]  W. K. Newey and D. McFadden, “Large sample estimation and hypothesis
testing,” \emph{Handbook of Econometrics}, vol. 4, pp. 2111–2245,
1994.
\item[{{[}2{]}}]  K. Hornik, M. Stinchcombe, and H.White, “Multilayer feedforward
networks are universal approximators,” \emph{Neural Netw.}, vol. 2,
no. 5, pp. 359–366, 1989.
\item[{{[}3{]}}]  B. Trendelkamp-Schroer and F. Noé, “Efficient estimation of rare-event
kinetics,” \emph{Phys. Rev. X}, vol. 6, pp. 011009, 2016.
\item[{{[}4{]}}]  M. K. Scherer, B. Trendelkamp-Schroer, F. Paul, G. Pérez-Hernández,
M. Hoffmann, N. Plattner, C. Wehmeyer, J. -H. Prinz, and F. Noé, ``PyEMMA
2: A software package for estimation, validation, and analysis of
Markov models,'' \emph{J. Chem. Theory Comput.}, vol. 11, no. 11,
pp. 5525-5542, 2015.
\end{enumerate}

\end{document}